\PassOptionsToPackage{table}{xcolor}

\documentclass[sigconf]{acmart}

\AtBeginDocument{%
  }

\acmYear{2026}
\acmConference[DAI '26]
  {The Eighth International Conference on Distributed Artificial Intelligence}
  {November 29--December 2, 2026}
  {Hong Kong, China}
\acmBooktitle{The Eighth International Conference on Distributed Artificial
  Intelligence (DAI '26), November 29--December 2, 2026, Hong Kong, China}

\setcopyright{none}
\copyrightyear{}
\acmDOI{}
\acmISBN{}

\usepackage{mathtools}
\usepackage{multirow}
\usepackage{enumitem}
\usepackage{wrapfig}
\usepackage{algorithm}
\usepackage{algpseudocode}
\usepackage{arydshln}

\usepackage[capitalize,noabbrev]{cleveref}

\definecolor{DeltaGreen}{RGB}{34,139,34}
\definecolor{DeltaRed}{RGB}{178,34,34}
\newcommand{\good}[1]{\textcolor{DeltaGreen}{#1}}
\newcommand{\bad}[1]{\textcolor{DeltaRed}{#1}}

\newcommand{\methodname}{ExecutionRAG}

\begin{document}

\title{Beyond Retrieval Failure: Understanding Why LLMs Miss Evidence in Multi-Hop RAG}

\author{Jiaqi Wei}
\authornote{Both authors contributed equally to this research.}
\affiliation{%
  \institution{Zhejiang University}
  \city{Hangzhou}
  \country{China}}
\affiliation{%
  \institution{Shanghai AI Laboratory}
  \city{Shanghai}
  \country{China}}
\email{jiaqi.wei@zju.edu.cn}

\author{Wenxuan Huang}
\authornotemark[1]
\affiliation{%
  \institution{Fudan University}
  \city{Shanghai}
  \country{China}}
\email{winshion0819@gmail.com}

\author{Chengdong Xu}
\affiliation{%
  \institution{Sun Yat-sen University}
  \city{Guangzhou}
  \country{China}}
\email{codepoet2077@gmail.com}

\author{Xiang Zhang}
\affiliation{%
  \institution{University of British Columbia}
  \city{Vancouver}
  \country{Canada}}
\email{xzhang23@ualberta.ca}

\author{Wanli Ouyang}
\affiliation{%
  \institution{The Chinese University of Hong Kong}
  \city{Hong Kong SAR}
  \country{China}}
\email{wlouyang@ie.cuhk.edu.hk}

\renewcommand{\shortauthors}{Wei et al.}

\begin{abstract}
Multi-hop retrieval-augmented generation (RAG) requires an agent to
retrieve evidence for a sequence of dependent sub-questions. When a
wrong answer lacks a required passage, the passage may be inaccessible
to the retriever, or the agent may have failed to issue a query that
would retrieve it. Standard retrieval evaluations conflate these two
causes. We study how much apparent retrieval failure is instead caused
by \emph{search drift}: the search no longer tracks an unanswered
sub-question. Holding the retriever, corpus, and generation model fixed,
we introduce \methodname, a training-free requirement register that
records open sub-questions and their supporting evidence. The register
changes search guidance but not evidence access, so each gold passage it
recovers certifies that the passage was reachable by the original
retriever.
On 1{,}000 MuSiQue questions, the number of wrong answers with
unretrieved required evidence falls from 382 to 234 with
all-MiniLM-L6-v2 and from 383 to 274 with Qwen3-Embedding-0.6B. These
counts correspond to
69\% to 44\% and 66\% to 48\% of each system's wrong answers,
respectively. Crucially, evidence recovery does not produce a
proportional reduction in total error: premature resolutions and
abstentions increase as missing-evidence failures decrease. The central
finding is therefore a shift in the remaining bottleneck, from guiding
search toward required evidence to verifying that the retrieved
evidence is sufficient.
\end{abstract}

\ccsdesc[500]{Information systems~Retrieval models and ranking}
\ccsdesc[500]{Information systems~Question answering}
\ccsdesc[300]{Computing methodologies~Natural language processing}

\keywords{Retrieval-augmented generation, multi-hop question answering,
  agentic retrieval, search drift, failure analysis}

\maketitle

\section{Introduction}

Retrieval-augmented generation (RAG)~\citep{lewis2020retrieval,gao2023retrieval}
grounds language-model answers in retrieved evidence. This is especially
important for multi-hop questions, whose answers depend on several
documents. Consider the question \emph{``Who married the publisher of
the abolitionist newspaper The North Star?''} Answering it requires two
linked facts: Frederick Douglass published \emph{The North Star}, and
Helen Pitts Douglass was his wife. A retrieval-augmented agent may
search for ``The North Star, publisher, married,'' anchor on William
Cooper Nell, who wrote for rather than published the paper, and spend
the remaining budget on Nell's marriage. Both required facts remain in
the corpus. The failure is not evidence access; it is the loss of the
intermediate entity that should guide the next query.

Most work on complex RAG focuses on evidence access. Methods organize a
corpus into graphs or hierarchies to make passages easier to
retrieve~\citep{edge2024local,sarthi2024raptor,guo2024lightrag}, while
agentic methods determine what to retrieve next. These approaches do
not address the example above when the retriever can already return the
needed passage but the search no longer represents the intermediate
target. On MuSiQue, maintaining the unresolved sub-questions across
steps, without changing the retriever or corpus, reduces the share of
wrong answers that lack a required passage from 382 to 234 cases under
one retriever and from 383 to 274 under another. This result motivates a
different question: how much apparent retrieval failure can be reduced
by improving search guidance without changing retriever coverage?

We call this failure \emph{search drift}. During a multi-hop search, the
agent may discard an unanswered sub-question, anchor on the wrong
intermediate entity, or keep querying for facts it has already
found~\citep{joren2024sufficient,ru2024ragchecker,kortukov2024contextmemory}.
Search drift appears as a retrieval symptom because a required passage
is absent from the trace, even though it is present in the corpus. It is
also the dominant failure category for a strong agentic baseline on
MuSiQue: 66--69\% of wrong answers omit at least one required passage,
and this share increases from 62\% at two hops to 81\% at four hops.

\methodname\ implements this measurement with an explicit requirement
register. The register is a list of the question's unresolved
sub-questions, initialized from the question and seed evidence rather
than from the question alone. Each entry records its state and the
passages that support it. Open entries guide subsequent search; an entry
is retired only when later evidence contradicts it; and the agent
answers only after no entry remains open. The language model still
decides whether evidence resolves an entry. Two deterministic rules
preserve the record by preventing open entries from silently disappearing
and by retaining all previously recorded evidence.

The main result is not simply that the register improves accuracy, but
that it reveals what happens after more evidence is recovered.
Missing-evidence failures decrease because the agent continues to search
for unresolved entries, yet premature resolutions and abstentions
increase, so total error falls by a much smaller amount. Because each
resolution is explicit, we can compare it with gold supporting evidence
and identify cases in which the agent answers before the evidence is
sufficient. The register therefore exposes a transition from a search
guidance problem to an evidence verification problem.

\textbf{In summary, this paper makes the following contributions:}
\begin{itemize}[leftmargin=*, topsep=2pt, itemsep=0pt]
    \item We distinguish two causes of missing required evidence:
    evidence that the fixed retriever cannot reach and evidence that the
    search never requests. This distinction turns evidence recovery
    under fixed retrieval into a diagnostic of search drift.
    \item We show a substantial controlled reduction in missing-evidence
    failures. On MuSiQue, the count decreases from 382 to 234 with
    all-MiniLM-L6-v2 and from 383 to 274 with Qwen3-Embedding-0.6B.
    However, premature resolutions and abstentions increase, making
    evidence verification the principal remaining bottleneck.
    \item We introduce \methodname, a training-free requirement register
    that provides the controlled intervention. It tracks unresolved
    sub-questions and their supporting passages while leaving the
    retriever, corpus, and generation-model parameters unchanged.
\end{itemize}

\begin{figure*}[!t]
    \centering
    \includegraphics[width=1\linewidth]{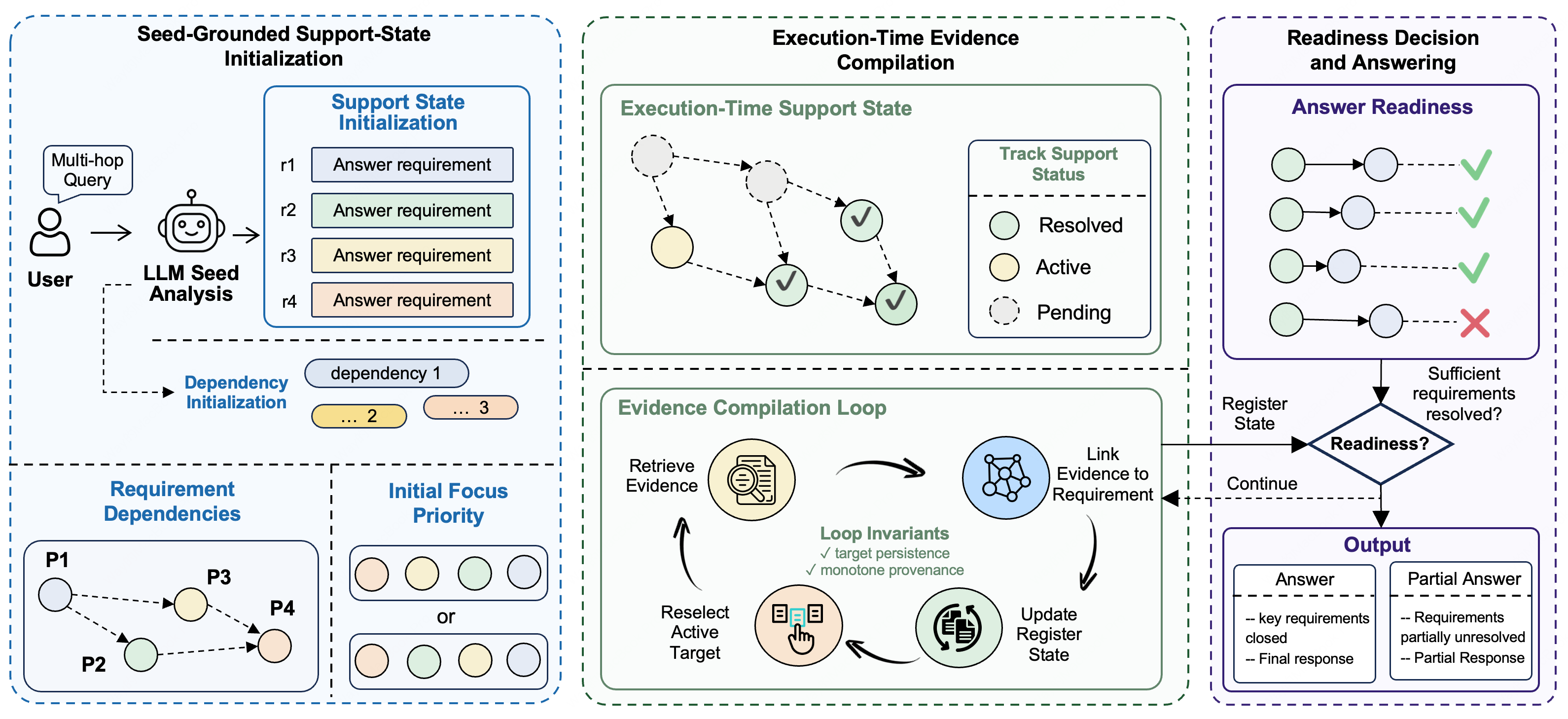}
    \Description{Three-stage schematic of ExecutionRAG: a seed retrieval builds a register of the question's sub-questions; each step selects an open item, searches for it, and marks it answered or retired; the agent answers once no item is still open.}
    \caption{\textbf{Overview of {\methodname}.}
    The agent keeps a requirement register of the question's open sub-questions and works through it in three stages:
    \textbf{(1)} build the register from the question and an initial (seed) retrieval, so its items are grounded in evidence the corpus can actually return;
    \textbf{(2)} at each step, select an open item, retrieve evidence,
    and update the register. The agent marks an item answered when a
    passage supports it and records that passage, or drops the item if
    later evidence contradicts it;
    and \textbf{(3)} answer only once no item is still open, or return a bounded partial answer when the search budget runs out.}
    \label{fig:method}
\end{figure*}

\section{Problem Setup: What ``Retrieval Failure'' Hides}
\label{sec:overview}

Given a query $q$ and a corpus $\mathcal{K}$, conventional RAG
retrieves context $\mathcal{C}$ and generates an answer in one shot,
$\mathcal{C}=R(q;\mathcal{K})$ and $a=G(q,\mathcal{C})$, treating the
retrieved passages as a single context.
For a multi-hop question, the challenge is not only whether the right
passages are retrieved, but also whether the agent retains which
sub-questions are resolved and which remain open as the search proceeds.

\subsection{Two Causes Inside One Symptom}
\label{sec:decomposition}

Multi-hop benchmarks let us say precisely what went wrong.
For a question $q$, let $P^\star(q)$ be its \emph{gold supporting
passages}: the minimal evidence a correct answer needs.
We call a wrong answer an \emph{apparent retrieval failure} when some
$p \in P^\star(q)$ was never retrieved.
Read literally, the name blames the retriever.
But with $R$ and $\mathcal{K}$ held fixed, the event splits into two
\emph{disjoint} causes: \textbf{unreachable}, no query the agent could
have issued would have made $R$ return $p$; and \textbf{unasked}, some
query would have returned $p$, but the agent never issued it.
Only the first is a retrieval problem; the second is \emph{search
drift}, where the evidence was within reach the whole time and the
search stopped asking for it.
The two are indistinguishable in a trace, because both end with the
same observable, namely a missing passage. The category is therefore
routinely attributed to the retriever.

\subsection{What a Fixed Retriever Identifies}
\label{sec:identification}

The unasked share is not directly observable: deciding that \emph{no}
query would have worked means quantifying over a query space we cannot
enumerate.
One direction, however, is decidable.
If an intervention leaves $R$ and $\mathcal{K}$ untouched and the
needed passage is nonetheless retrieved, then a query returning it
existed, so that failure was unasked.
Recovering evidence \emph{certifies reachability}; failing to recover
it proves nothing.
Holding the retriever fixed is therefore necessary for the inference.

This argument identifies reachability at the level of an individual
recovered passage. At the aggregate level, however, a reduction in the
frequency of missing-evidence failures does not by itself identify which
baseline failures were converted, because the two systems can fail on
different questions. We therefore use the controlled reduction as
evidence of a recoverable component of apparent retrieval failure, not
as a numerical lower bound on the total prevalence of search drift. The
minimal intervention keeps this interpretation specific to search
guidance without claiming that every remaining missing passage is
unreachable.

\subsection{The Probe: An Explicit Requirement Register}
\label{sec:probe}

We therefore need an intervention that keeps the open sub-questions in
view while changing nothing about retrieval: an explicit \emph{requirement register}, in effect a checklist the agent must clear before it answers.
For a query $q$, let $\mathcal{D}(q)=\{d_1,\dots,d_m\}$ be the
question's answer requirements, namely the sub-questions the answer
must satisfy (e.g., ``who is the publisher'' and ``who did that person
marry'').
At step $t$ the register records, for each item, its status $z_i^{(t)}$,
a short evidence note $e_i^{(t)}$, and the passages $P_i^{(t)}$ that
support it:
\begin{equation}
\mathcal{S}_t=\{(d_i,z_i^{(t)},e_i^{(t)},P_i^{(t)})\}_{i=1}^{m}.
\end{equation}
It says, at a glance, what is answered and what is still open, and it
tells the agent what to search for next: an item that is still open,
rather than repeatedly searching from $q$ alone.
Because $P_i^{(t)}$ names the passage the agent \emph{believed} settled
$d_i$, it can be checked against $P^\star(q)$ afterwards, which is what
turns the register from a control structure into an instrument
(Section~\ref{sec:error}).

\section{Method}
\label{sec:method}

{\methodname} is deliberately the smallest intervention that satisfies
Section~\ref{sec:identification}: it changes what the agent searches
\emph{for}, and nothing about how retrieval works.
The retriever, the corpus, the tool interface, and the generator are
untouched, so any evidence it recovers was evidence the same retriever
could already reach.

\subsection{What the Register Records}
\label{sec:register}

{\methodname} maintains one entry per item, holding its status, a short evidence
note, and the passages that support it.
Each item's status takes one of four values:
\begin{equation}
z_i^{(t)} \in \{\texttt{pending},\texttt{active},\texttt{resolved},\texttt{retired}\}.
\end{equation}
\texttt{pending} means the item has been identified but is not the
current search target;
\texttt{active} is the item currently guiding the search;
\texttt{resolved} means enough evidence has been found to answer it;
and \texttt{retired} means it has been dropped, and only with an
explicit reason.

We keep the status set deliberately small.
Partial progress lives in the evidence note $e_i^{(t)}$ and the
supporting passages $P_i^{(t)}$, not in extra labels, so an item stays
open until its evidence is actually enough.
The status says \emph{whether} an item is answered; the note and
passages say \emph{how}.
The four values are also what make the agent's judgment legible: the
step from \texttt{active} to \texttt{resolved} is the moment it
declares an item settled. Recording this decision as an event, rather
than leaving it implicit in free-form reasoning, allows us to compare it
with gold evidence and determine whether it was premature.

\subsection{Building the Register from Seed Evidence}
\label{sec:init}

The register should not be written from the question alone.
Splitting a question in the abstract tends to produce items that are vague, redundant, or ask for things the corpus cannot answer.
So {\methodname} first runs one \emph{seed} retrieval and builds the register from what comes back:
\begin{equation}
\mathcal{C}^{(0)} = R_{\text{seed}}(q;\mathcal{K}).
\end{equation}
The model reads $(q,\mathcal{C}^{(0)})$ and writes the initial register $\mathcal{S}_0$, so its items are shaped by the question but grounded in evidence the corpus can actually return.
One item is set \texttt{active} as the first search target and the rest start \texttt{pending}, with empty notes:
\begin{equation}
\begin{gathered}
z_{b_0}^{(0)}=\texttt{active}, \qquad
z_i^{(0)}=\texttt{pending}\ (i\neq b_0),\\
e_i^{(0)}=\varnothing, \qquad
P_i^{(0)}=\varnothing.
\end{gathered}
\end{equation}
This gives the agent one concrete thing to search for while keeping the other sub-questions on the list for later.

\subsection{Updating the Register and Steering the Search}
\label{sec:transition}

Starting from $\mathcal{S}_0$, {\methodname} repeats a simple loop: pick an open item, search for it, and update that item.

\paragraph{Searching for one open item}
At step $t$, {\methodname} picks an open item $b_t$ from $\mathcal{S}_{t-1}$ and builds a query for it:
\begin{equation}
u_t = \textsc{Compose}(q, d_{b_t}, e_{b_t}^{(t-1)}).
\end{equation}
The original question $q$ keeps the overall goal in view, $d_{b_t}$ says which item is being chased, and the note $e_{b_t}^{(t-1)}$ points the search at what is still missing for that item.
The retriever returns new evidence $\Delta\mathcal{C}_t$; repeated identical queries are cached so the agent does not loop.

\paragraph{Updating only the items it touches}
Given $\Delta\mathcal{C}_t$, the register is updated as
\begin{equation}
\mathcal{S}_t = U(\mathcal{S}_{t-1}, \Delta\mathcal{C}_t).
\end{equation}
The update rewrites the notes for the items the new evidence bears on, adds the supporting passages, and moves an item to \texttt{resolved} only when there is enough support, or to \texttt{retired} only with an explicit reason.
Items the new evidence does not touch are left unchanged.
In effect, each retrieval is turned into progress on one open item, rather than piling up more unsorted context.

\subsection{When to Answer, and Two Bookkeeping Rules}
\label{sec:answering}

\paragraph{Answering only when the register is clear}
After each update, {\methodname} checks whether any item is still open.
If none is, it writes the final answer from the register and its supporting passages; otherwise it keeps searching for the items that remain open.
If the search budget runs out first, it returns a bounded partial answer that says plainly which items are still unsupported.

\paragraph{Showing only what changed}
To keep the context short, {\methodname} does not replay the whole register after every step.
It shows the agent only a \emph{progress delta}: the items whose status or note changed from $\mathcal{S}_{t-1}$ to $\mathcal{S}_t$.
This keeps the signal the agent needs for its next move without re-reading the full list each time.

\paragraph{Two rules that run outside the model}
Because the register is updated by multiple LLM calls, {\methodname}
enforces two deterministic rules to preserve its state. First, an open
item can leave the list only through an explicit transition to
\texttt{retired}:
\begin{equation}
\mathrm{Req}(\mathcal{S}_t)
\supseteq
\mathrm{Req}(\mathcal{S}_{t-1})
\setminus
\{d_k \mid z_k^{(t)}=\texttt{retired}\}.
\end{equation}
Second, supporting evidence is appended rather than overwritten:
\begin{equation}
P_i^{(t)}=P_i^{(t-1)}\cup \Delta P_i^{(t)}.
\end{equation}
A later update may revise an item's note, but it cannot erase the record
of passages that previously supported it. These rules preserve the
requirements and their evidence trail; they do not guarantee that the
language model resolves an item correctly. Keeping this decision
explicit allows us to evaluate it against gold evidence
(Section~\ref{sec:error}).

\section{Experiments}

\subsection{Experimental Questions}

Section~\ref{sec:identification} defines the controlled comparison:
holding the retriever and corpus fixed, we measure how much the
frequency of missing-evidence failures changes when only search guidance
is modified. The experiments first quantify this reduction
and then examine whether recovering evidence reduces total error or
reveals a different bottleneck.
The framing also makes predictions that could fail.
If the recovered evidence were an artifact of the register writing
longer answers that favor the judge, it would not improve recall of gold
evidence or the objective substring metric.
If the register keeps search on target without making the generator a
better verifier, the failures left behind should move from ``never
retrieved the evidence'' toward answering too early or abstaining.
And if the effect were really about reading more text, it should vanish
where the register reads no more than the baseline.

Accordingly, we ask:
\begin{itemize}[leftmargin=1.5em, topsep=2pt, itemsep=2pt]
    \item \textbf{RQ1}: Under a fixed retriever, how much does search
      guidance reduce failures caused by missing evidence?
    \item \textbf{RQ2}: How does the failure distribution change after
      more required evidence is recovered?
    \item \textbf{RQ3}: Does the reduction persist with a different
      embedding model?
    \item \textbf{RQ4}: Does the recovered evidence improve answer
      accuracy, particularly when sub-questions must be combined across
      documents?
    \item \textbf{RQ5}: Is the gain explained by retrieving
      more context?
    \item \textbf{RQ6}: Do seed grounding and iterative updates both
      contribute to the result?
\end{itemize}

\subsection{Experimental Setup}

\paragraph{Datasets}
We evaluate on three standard multi-hop QA benchmarks: HotpotQA~\cite{hotpotqa}, MuSiQue~\cite{musi}, and 2WikiMultiHopQA (2Wiki)~\cite{2wiki}.
Following HippoRAG, we use the same retrieval corpora and sample 1,000 questions from each validation set as test queries.

\paragraph{Baselines}
We compare against three groups of baselines:
(1)~\emph{Direct inference}: GPT-4o-mini without retrieval;
(2)~\emph{Vanilla RAG}: top-$k$ retrieval with $k\in\{3,5,10\}$, CoT-augmented retrieval, and IRCoT~\cite{IRCOT};
(3)~\emph{Structured and agentic RAG}: KGP~\cite{KGP}, G-Retriever~\cite{G-retriever}, RAPTOR~\cite{raptor}, LightRAG~\cite{lightrag}, HippoRAG and HippoRAG~2~\cite{hipporag,hippo2}, GFM-RAG~\cite{GFM}, LinearRAG~\cite{linear}, LAG~\cite{xiao2025lag}, and A-RAG~\cite{du2026rag}.
This comparison spans direct answering, flat retrieval, corpus
structuring, and iterative retrieval control.

\paragraph{Metrics}
Following LinearRAG~\cite{linear}, we report \emph{Contain-Match Accuracy}, which checks whether the gold answer appears as a substring of the prediction, and \emph{LLM-Evaluation Accuracy}, which uses an LLM judge to assess semantic equivalence.

\paragraph{Implementation details}
Unless otherwise specified, all methods use GPT-4o-mini as the generation model.
For the main comparison, we use all-MiniLM-L6-v2 with top-$k=5$ retrieval, and additionally evaluate Qwen3-Embedding-0.6B as an alternative embedding backbone.
All main results use the full 1,000-question evaluation sets.
The ablation study is conducted on the full 1{,}000-question MuSiQue evaluation set.
Additional implementation details are provided in Appendix~\ref{app:impl}.

\subsection{RQ1: Reduction in Failures Caused by Missing Evidence}
\label{sec:error}

We measure the controlled reduction described in
Section~\ref{sec:identification} by assigning each wrong answer on
MuSiQue a single cause using the gold supporting paragraphs. The full
counts and percentages for both retrievers appear in
Appendix~\ref{app:error}.

\paragraph{How we label a failure}
Each wrong answer receives exactly one cause, assigned by a fixed
priority order so the categories cannot overlap: \emph{judge error}
(the substring metric counts the answer as right but the judge does
not), \emph{gave up} (the answer explicitly abstains), \emph{reasoning
error} (every supporting passage was retrieved and the answer is still
wrong, or, for {\methodname}, an item was resolved while supporting
evidence was still missing), and \emph{missing evidence} (a supporting
passage was never retrieved and no premature resolution occurred).
What makes this labelling possible at all is that MuSiQue ships, for
each question, the supporting paragraphs a correct answer minimally
needs: we match each supporting paragraph to its corpus chunk by a
40-character fingerprint (87.9\% of paragraphs match) and ask whether
the search ever retrieved it.
Because the two methods fail on different numbers of questions, every
share below is a percentage of that method's own failures.

\begin{figure*}[!t]
    \centering
    \includegraphics[width=0.9\linewidth]{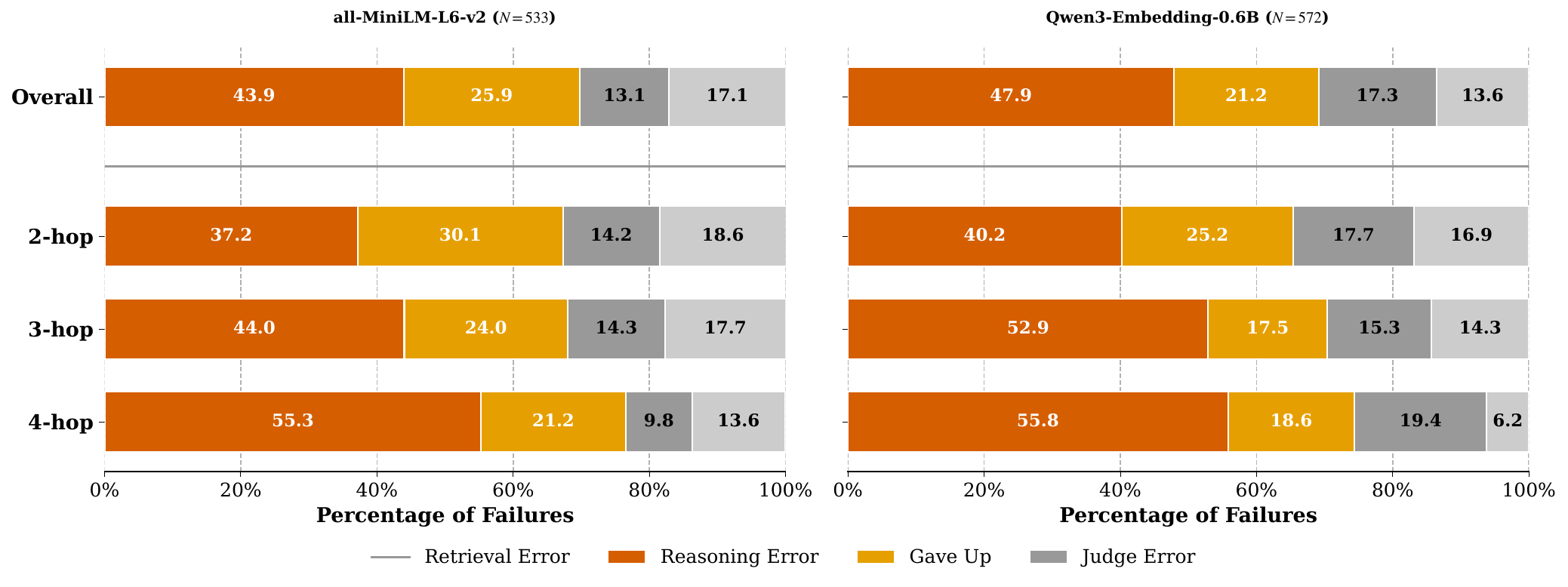}
    \Description{Stacked bars of wrong-answer causes on MuSiQue for A-RAG and ExecutionRAG, overall and split by hop count. The share that never retrieves a required passage decreases under ExecutionRAG while reasoning errors after retrieval and abstentions increase.}
    \caption{\textbf{Failure distribution on MuSiQue}, by failure type and hop count. The share of failures that never retrieve a required passage decreases after adding the register, while reasoning errors after retrieval and abstentions increase. For {\methodname}, failures caused by missing evidence remain most common at four hops.}
    \label{fig:error}
\end{figure*}

The largest change is a reduction in failures caused by missing
evidence: wrong answers for which a required passage was never
retrieved. With all-MiniLM-L6-v2, the count decreases from 382 for
A-RAG to 234 for {\methodname}; these values represent 69.3\% of 551
baseline failures and 43.9\% of 533 {\methodname} failures. With
Qwen3-Embedding-0.6B, the count decreases from 383 to 274, or from
66.4\% of 577 failures to 47.9\% of 572 failures. The same retriever is
used within each comparison, so the reductions show that search
guidance can recover a substantial portion of evidence that the
baseline search did not retrieve. They do not imply that every residual
missing passage is unreachable (Figure~\ref{fig:error}).

Recovering this evidence does not reduce total error proportionally.
Reasoning errors after retrieval increase from 9\% to 26\% with MiniLM
and from 8\% to 21\% with Qwen; these errors include cases in which the
register is resolved before its evidence is sufficient. Abstentions also
increase, from 7\% to 13\% and from 12\% to 17\%, respectively. Against
A-RAG, LLM-Evaluation Accuracy on MuSiQue rises from 44.9 to 50.2, a
smaller change than the reduction in missing-evidence failures. This
redistribution is the central result: once search guidance recovers more
required evidence, evidence verification and answer commitment become
the dominant sources of the remaining error.

\subsection{RQ2: Evidence Recovery Shifts the Remaining Failures}
\label{sec:residual}

\paragraph{Why reasoning errors become more visible}
Part of this increase reflects observability rather than a change in
the model's reasoning alone. A premature resolution is an explicit
register event and can therefore be assigned to reasoning error. A-RAG
has no analogous event; when it answers from incomplete evidence, the
error is generally assigned to the missing passage instead. The
comparison may thus undercount premature answers by the baseline. The
register exposes this decision for evaluation; it does not establish
that the generation model reasons worse.

The register guides search but does not provide a reliable evidence
verifier: the language model still decides whether a requirement has
been resolved, and this decision can be incorrect. It also cannot make
genuinely difficult passages retrievable. Residual failures caused by
missing evidence for {\methodname} remain highest on four-hop questions, which
contain longer and more weakly connected evidence chains. In this
setting, the remaining bottlenecks are verifying retrieved evidence and
retrieving evidence for long chains.

\subsection{RQ3: The Reduction Persists Across Embedding Models}
\label{sec:backbone}

\begin{figure*}[!t]
    \centering
    \includegraphics[width=0.9\linewidth]{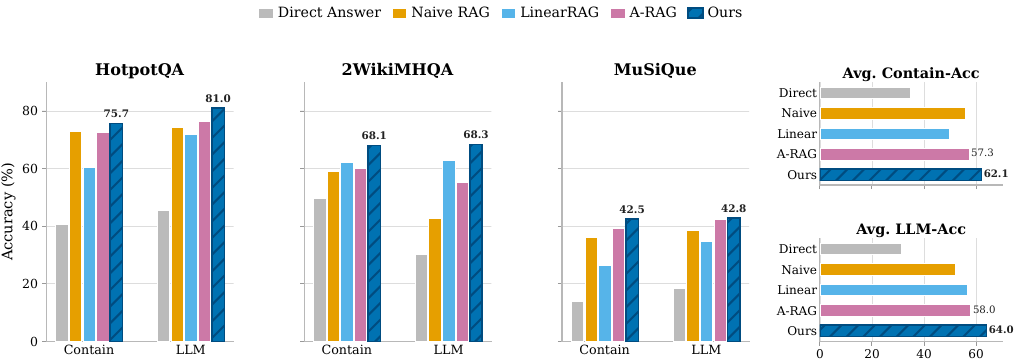}
    \Description{Bar charts of Contain-Match and LLM-Evaluation accuracy per benchmark, plus the cross-benchmark average, under the Qwen3-Embedding-0.6B retriever. ExecutionRAG stays above A-RAG on all three benchmarks.}
    \caption{\textbf{Performance under Qwen3-Embedding-0.6B.} Left: per-benchmark Contain-Match and LLM-Evaluation Accuracy. Right: average accuracy across benchmarks.}
    \label{fig:qwen}
\end{figure*}

Figure~\ref{fig:qwen} tests whether the performance pattern depends on
the embedding model. After replacing all-MiniLM-L6-v2 with
Qwen3-Embedding-0.6B,
{\methodname} outperforms A-RAG on all three benchmarks: 80.0, 68.3,
and 42.8 LLM-Evaluation Accuracy on HotpotQA, 2Wiki, and MuSiQue,
respectively, compared with 76.4, 55.2, and 42.3 for A-RAG. The largest
gain remains on 2Wiki ($+13.1$), while MuSiQue shows a smaller positive
gain ($+0.5$). The positive difference across all three benchmarks shows
that the effect is not specific to one embedding model. The larger gain
on 2Wiki is also consistent with the proposed mechanism because its
questions require dependent sub-questions to be maintained across
multiple retrieval steps.

\begin{table*}[!t]
    \centering
    \small
    \renewcommand{\arraystretch}{1.1}
    \setlength{\tabcolsep}{9pt}

    \caption{\textbf{Main results on three multi-hop QA benchmarks.}
    Bold marks the best \emph{Contain-Match} result per benchmark; the
    grey columns report \emph{LLM-Evaluation Accuracy} for comparability
    with prior work. These answer-level metrics complement the primary
    failure analysis by showing whether recovered evidence improves the
    final prediction.
    The lower block is the matched comparison our attribution rests on:
    A-RAG shares {\methodname}'s generator, tools, and corpora, so it
    is {\methodname} \emph{without} the register.
    Gains concentrate where sub-questions must be combined across
    documents; HotpotQA is a tie.}
    \label{tab:main_results}

    \begin{tabular}{@{} l ccc @{\hspace{14pt}} ccc @{}}
        \toprule
        \multirow{2}{*}{\textbf{Method}}
          & \multicolumn{3}{c}{\textbf{Contain-Match}}
          & \multicolumn{3}{c}{\textcolor{gray}{LLM-Evaluation}} \\
        \cmidrule(lr){2-4} \cmidrule(lr){5-7}
        & Hotpot & 2Wiki & MuSiQue
        & \textcolor{gray}{Hotpot} & \textcolor{gray}{2Wiki} & \textcolor{gray}{MuSiQue} \\
        \midrule

        \rowcolor{gray!10} \multicolumn{7}{@{}l}{\textit{Direct Zero-shot LLM Inference}} \\
        GPT-4o-mini~\cite{gpt} & 30.4 & 29.0 & 7.8 & \textcolor{gray}{34.2} & \textcolor{gray}{28.6} & \textcolor{gray}{10.1} \\

        \rowcolor{gray!10} \multicolumn{7}{@{}l}{\textit{Vanilla Retrieval-Augmented Generation}} \\
        Retrieval (Top-3) & 52.1 & 45.1 & 23.4 & \textcolor{gray}{55.1} & \textcolor{gray}{43.1} & \textcolor{gray}{27.1} \\
        Retrieval (Top-5) & 54.6 & 46.6 & 25.6 & \textcolor{gray}{56.8} & \textcolor{gray}{45.3} & \textcolor{gray}{29.0} \\
        Retrieval (Top-10) & 56.0 & 48.7 & 26.7 & \textcolor{gray}{58.6} & \textcolor{gray}{45.8} & \textcolor{gray}{31.2} \\
        CoT (Top-5)~\cite{cot} & 55.1 & 48.7 & 27.1 & \textcolor{gray}{57.1} & \textcolor{gray}{45.9} & \textcolor{gray}{30.7} \\
        IRCoT (Top-5)~\cite{IRCOT} & 58.4 & 53.0 & 22.6 & \textcolor{gray}{59.6} & \textcolor{gray}{36.8} & \textcolor{gray}{26.1} \\

        \rowcolor{gray!10} \multicolumn{7}{@{}l}{\textit{Graph and Agentic Retrieval-Augmented Generation}} \\
        KGP~\cite{KGP} & 56.2 & 52.2 & 30.5 & \textcolor{gray}{57.1} & \textcolor{gray}{33.9} & \textcolor{gray}{27.3} \\
        G-retriever~\cite{G-retriever} & 41.3 & 47.8 & 14.1 & \textcolor{gray}{40.9} & \textcolor{gray}{25.7} & \textcolor{gray}{15.6} \\
        RAPTOR~\cite{raptor} & 58.1 & 60.6 & 32.2 & \textcolor{gray}{55.3} & \textcolor{gray}{43.9} & \textcolor{gray}{29.7} \\
        LightRAG~\cite{lightrag} & 61.5 & 54.4 & 27.7 & \textcolor{gray}{60.5} & \textcolor{gray}{38.0} & \textcolor{gray}{28.3} \\
        HippoRAG (single-step)~\cite{hipporag} & 55.2 & 63.7 & 31.4 & \textcolor{gray}{57.9} & \textcolor{gray}{57.5} & \textcolor{gray}{30.1} \\
        HippoRAG (multi-step)~\cite{hipporag} & 61.1 & 66.4 & 34.0 & \textcolor{gray}{63.6} & \textcolor{gray}{62.4} & \textcolor{gray}{31.8} \\
        GFM-RAG (single-step)~\cite{GFM} & 61.4 & 66.2 & 29.3 & \textcolor{gray}{64.8} & \textcolor{gray}{61.1} & \textcolor{gray}{32.6} \\
        GFM-RAG (multi-step)~\cite{GFM} & 63.4 & 69.5 & 31.5 & \textcolor{gray}{65.5} & \textcolor{gray}{63.2} & \textcolor{gray}{35.5} \\
        HippoRAG 2~\cite{hippo2} & 61.2 & 62.0 & 34.5 & \textcolor{gray}{64.3} & \textcolor{gray}{58.8} & \textcolor{gray}{35.6} \\
        LinearRAG~\cite{linear} & 64.2 & 69.5 & 32.8 & \textcolor{gray}{64.9} & \textcolor{gray}{62.6} & \textcolor{gray}{36.9} \\
        LAG~\cite{xiao2025lag} & 68.5 & 71.9 & 42.8 & \textcolor{gray}{69.7} & \textcolor{gray}{64.0} & \textcolor{gray}{44.3} \\

        \midrule
        \rowcolor{gray!10} \multicolumn{7}{@{}l}{\textit{Matched condition: same generator, retrieval tools, and corpora}} \\
        A-RAG~\cite{du2026rag}\quad\emph{(= ours w/o register)} & \textbf{77.3} & 66.7 & 42.0 & \textcolor{gray}{78.6} & \textcolor{gray}{59.8} & \textcolor{gray}{44.9} \\
        \rowcolor{blue!5} \textbf{{\methodname} (Ours)} & 77.0 & \textbf{73.0} & \textbf{49.1} & \textcolor{gray}{79.8} & \textcolor{gray}{74.3} & \textcolor{gray}{50.2} \\
        \quad\emph{$\Delta$ vs.\ matched} & \bad{$-0.3$} & \good{$+6.3$} & \good{$+7.1$} & \textcolor{gray}{$+1.2$} & \textcolor{gray}{$+14.5$} & \textcolor{gray}{$+5.3$} \\
        \specialrule{\heavyrulewidth}{0pt}{\belowrulesep}
    \end{tabular}
\end{table*}

\subsection{RQ4: Answer Accuracy}
\label{sec:overall}

Table~\ref{tab:main_results} reports the main comparison. We first
consider Contain-Match Accuracy because it does not rely on an LLM
judge. On MuSiQue, {\methodname} reaches 49.1, a 6.3-point improvement
over LAG (42.8). On 2Wiki, it reaches 73.0, a 1.1-point improvement
over LAG (71.9). On HotpotQA, it reaches 77.0, which is essentially tied
with A-RAG (77.3). LLM-Evaluation Accuracy follows the same pattern and
is highest for {\methodname} on all three benchmarks: 79.8, 74.3, and
50.2 on HotpotQA, 2Wiki, and MuSiQue, respectively.

The differences are consistent with the proposed mechanism. The
register has little effect on HotpotQA, whose questions are mostly
two-hop and are often answerable from a single comparison. It improves
performance on 2Wiki and MuSiQue, where the agent must maintain several
dependent sub-questions across documents.

A-RAG is the matched baseline for the failure analysis. It uses the same
generation model, retrieval tools, and corpora as {\methodname}; the
requirement register is the added component. The difference between
A-RAG and {\methodname} in Sections~\ref{sec:error}--\ref{sec:backbone}
therefore cannot be attributed to a different retriever or generation
backbone. The register adds bookkeeping calls, but these calls alone do
not explain the gain: on 2Wiki, the largest improvement occurs with
comparable or fewer retrieved tokens (Section~\ref{sec:efficiency}).

\paragraph{Where the gains concentrate}
\label{sec:category}

Figure~\ref{fig:category} in Appendix~\ref{app:category} reports
category-level results. Under both embedding models, questions with
longer or less direct evidence chains remain more difficult. The
register helps most when required evidence is retrievable but must be
tracked across several steps. On MuSiQue, accuracy decreases with hop
count. On HotpotQA, Comparison questions are easier than Bridge
questions; on 2Wiki, Inference questions are hardest and Comparison
questions are easiest. The register improves overall accuracy but does
not remove the difficulty of long evidence chains.

\begin{figure*}[!t]
    \centering
    \includegraphics[width=0.96\linewidth]{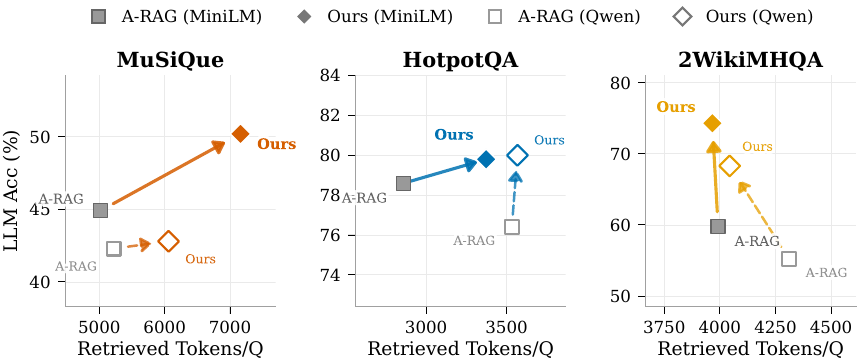}
    \Description{Pareto scatter of LLM-Evaluation accuracy against retrieved tokens per question for ExecutionRAG and A-RAG under two embedding backbones, with ablation variants below.}
    \caption{\textbf{Accuracy--efficiency trade-off}. Pareto frontiers of LLM-Evaluation Accuracy vs.~retrieved tokens per question for {\methodname} and A-RAG under two embedding backbones. Bottom row: ablation variants; bars show LLM-Acc and the line shows average retrieved tokens per question.}
    \label{fig:efficiency}
\end{figure*}

\subsection{RQ5: Retrieved Context and Accuracy}
\label{sec:efficiency}

Figure~\ref{fig:efficiency} and Table~\ref{tab:ablation} test whether
the gains can be explained by retrieving more context. On MuSiQue and
HotpotQA, {\methodname} reads more tokens than A-RAG because it builds
and updates the register. On 2Wiki, however, it uses comparable or fewer
retrieved tokens while obtaining its largest accuracy improvement. Within
{\methodname}, removing register components reduces both token use and
accuracy. The result is consistent with the register improving the
targeting of retrieval rather than simply increasing the amount of
retrieved context.

\paragraph{What the register costs}
The register increases the cost of some searches. On MuSiQue, it reads
7{,}161 retrieved tokens per question, compared with 5{,}015 for A-RAG;
on HotpotQA, it reads 3{,}373 compared with 2{,}860. On 2Wiki, token use
is comparable at 3{,}966 and 3{,}992, respectively. The corresponding
API costs are \$0.0064 versus \$0.0034 on MuSiQue and \$0.0039 versus
\$0.0027 on 2Wiki. Search is limited to fifteen retrieval loops and a
cumulative context of 128{,}000 tokens. The agent answers when no
requirement remains open, so the register shortens some trajectories and
lengthens others; the case studies in Appendix~\ref{app:case} complete
in two and six loops, whereas A-RAG uses five and thirteen. Appendix~
\ref{app:efficiency} provides the complete breakdown.

\begin{table}[t]
    \centering
    \caption{\textbf{Ablation study on MuSiQue.} LLM-Evaluation and Contain-Match Accuracy, under the default setup.}
    \label{tab:ablation}
    \small{%
      \begin{tabular}{lcc}
      \toprule
      \textbf{Method} & \textbf{LLM-Acc} & \textbf{Contain-Acc} \\
      \midrule
      \textbf{\methodname} & \textbf{50.2} & \textbf{49.1} \\
      w/o Register Init. & 48.7 & 46.2 \\
      w/o Progress Register & 48.3 & 48.5 \\
      \bottomrule
      \end{tabular}
    }
\end{table}

\subsection{RQ6: Effects of Seed Grounding and Iterative Updates}
\label{sec:ablation}

Table~\ref{tab:ablation} isolates the contributions of the register's
components. Building the register from the question alone rather than
from seed evidence lowers LLM-Acc from 50.2 to 48.7, indicating that
grounding requirements in retrievable evidence is beneficial. Removing
iterative updates and constructing the register only once lowers
LLM-Acc further to 48.3. Both seed grounding and iterative updates
contribute to the observed gain.

\section{Related Work}
\label{app:related}

\paragraph{Corpus structure for RAG}
A major line of work improves multi-hop QA by structuring corpus
knowledge to improve evidence access.
GraphRAG~\citep{edge2024local} builds entity-relation graphs with community summaries;
RAPTOR~\citep{sarthi2024raptor} organizes documents into hierarchical retrieval trees;
LightRAG~\citep{guo2024lightrag} combines graph-augmented indexing with efficient retrieval;
HippoRAG and HippoRAG~2~\citep{jimenez2024hipporag,hipporag2} use graph-based retrieval with Personalized PageRank;
and LinearRAG~\citep{linear}, GFM-RAG~\citep{GFM}, and StructRAG~\citep{li2024structrag} further explore graph construction, graph foundation models, and hybrid structurization.
These methods improve how dispersed evidence is represented and
retrieved before answering. {\methodname} is complementary: it does not
restructure the corpus, but maintains an explicit register of unresolved
sub-questions during search.

\paragraph{Iterative and agentic RAG}
A related line of work improves retrieval control during inference.
IRCoT~\citep{IRCOT} and FLARE~\citep{jiang2023active} interleave reasoning with retrieval;
Self-RAG~\citep{asai2023selfraglearningretrievegenerate} uses reflection tokens to critique retrieved passages;
and recent agentic methods expose richer retrieval policies through tool use, query reformulation, and multi-step search.
A-RAG~\citep{du2026rag} replaces fixed pipelines with model-driven hierarchical retrieval interfaces, while LAG~\citep{xiao2025lag} and LogicRAG~\citep{chen2026you} decompose questions into subproblems, infer dependencies, and sequence retrieval accordingly.
This view is also related to work that formulates language-model
inference as search~\citep{wei2026llm}, analyzes how chain-of-thought
prompting interacts with the answer space~\citep{zhang2025does}, learns
when a model should disclose its reasoning~\citep{wei2026think}, learns
execution-time workflows for multi-agent systems~\citep{xu2026evomas}, or
scales multi-agent processing over many documents at test
time~\citep{cao2025multi2}.
These methods mainly optimize retrieval control: what to retrieve next
and in what order. {\methodname} does not introduce a new retrieval
policy. Instead, it maintains an explicit register of unresolved
sub-questions that guides subsequent search and determines when the
agent can answer. Even with adaptive retrieval, an agent can discard an
unresolved sub-question, overwrite earlier progress, or answer from
partial evidence. The register preserves these requirements throughout
the search.
It is also narrower than agent-memory work that retrieves over a
persistent episodic record of an agent's past
decisions~\citep{xu2026agent}: the register tracks the open requirements
of the \emph{current} question only, and is discarded once the answer is
given.

\paragraph{Long executions outside question answering}
The same pressure to keep intermediate state explicit appears in domains
where an agent must execute a long chain before it can commit to an
answer. Multi-agent drafting protocols compress intermediate
reasoning~\citep{li2026draftrl}, and retrieval-augmented multi-agent
pipelines are built for tasks whose verdict depends on evidence gathered
in several steps~\citep{li2026retrievalaugmentedfakenews}.
Software-engineering agents are trained against execution feedback and
process-level reward precisely because their trajectories are too long to
supervise end to
end~\citep{li2026boostaprboostingautomatedprogram,li2026counterfactualrolloutreplayforkable,tan2026evocurl}.
There too the difficulty is less the first attempt than checking what the
agent has actually done. The register is the minimal instance of that bookkeeping
for retrieval: it records the requirements a trajectory still owes, and
nothing else.

\paragraph{Sufficiency signals, reading notes, and failure analysis}
Several methods track information gaps or judge evidence sufficiency
during RAG. Self-RAG~\citep{asai2023selfraglearningretrievegenerate}
critiques passages with reflection tokens~\citep{zhang2025tokenization}. Related work adds corrective
retrieval (CRAG~\citep{yan2024corrective}), sequential reading notes
(Chain-of-Note~\citep{yu2023chainofnote}), and structured
missing-information items with sufficiency checks
(S2G-RAG~\citep{li2026s2grag}, SURE-RAG~\citep{qiu2026surerag}).
Test-time critique and optimization provide another mechanism for
improving retrieved evidence and generated answers~\citep{weiretrieval}. Other
methods diagnose or repair agent trajectories after the fact, such as
Doctor-RAG~\citep{jiao2026doctorrag}. The requirement register is an
explicit list of sub-questions with states and evidence provenance. Our
contribution is empirical: by holding the retriever and generator fixed,
we measure how such a register changes the multi-hop failure
distribution. It reduces failures caused by missing evidence under a
fixed retriever and shifts residual errors toward reasoning after
retrieval, a distinction that prior end-to-end evaluations do not
isolate.

\paragraph{Naming the two causes behind one symptom}
Skill-RAG~\citep{wei2026skillrag} makes the closest claim to
Section~\ref{sec:decomposition} we know of: retrieval failures ``stem
not from the absence of relevant evidence but from an alignment gap
between the query and the evidence space,'' with an exit route for
cases it calls truly irreducible.
This distinguishes unreachable and unasked causes conceptually and
routes learned repair skills accordingly. It does not define the failure
event using gold evidence or quantify either share. We instead treat the
distinction as the measurement target. Outside retrieval, Binding
Drift~\citep{babu2026bindingdrift} separates drift from error propagation
using disjoint evaluation sets. We apply the same separation to a
missing passage by asking whether the search could have retrieved it.

\paragraph{Implications}
The results distinguish two directions for future work. Because the
recovered evidence was reachable by the same retriever, increasing index
coverage, improving embeddings, or expanding queries does not address
the observed failures associated with inadequate search guidance. Those
failures require a mechanism that preserves unresolved sub-questions.
Conversely,
the increase in premature resolution identifies evidence verification as
a separate target. The register makes the resolution decision explicit,
which enables a verifier to be trained or evaluated against it. Residual
failures caused by missing evidence concentrate on long, weakly connected chains,
where improving retrieval coverage remains necessary. These predictions
are testable: a verifier should improve answer accuracy without changing
retrieved evidence, whereas an intervention that improves only coverage
should not eliminate failures caused by unasked queries.

\section{Conclusion}

We study how much apparent retrieval failure in multi-hop RAG can be
reduced by search guidance while the retriever remains fixed. On
MuSiQue, the requirement register reduces wrong answers with
unretrieved required evidence from 382 to 234 cases under one retriever
and from 383 to 274 under another. This recovery does not translate
proportionally into answer accuracy: premature resolutions and
abstentions increase as missing-evidence failures decrease. The result
separates two bottlenecks that aggregate accuracy obscures. Maintaining
unresolved requirements helps the agent retrieve reachable evidence,
but converting that evidence into correct answers requires more reliable
verification after retrieval.

\begin{acks}
Supported by Shanghai Artificial Intelligence Laboratory.
\end{acks}

\bibliographystyle{ACM-Reference-Format}
\bibliography{main}


\begin{thebibliography}{54}


\ifx \showCODEN    \undefined \def \showCODEN     #1{\unskip}     \fi
\ifx \showISBNx    \undefined \def \showISBNx     #1{\unskip}     \fi
\ifx \showISBNxiii \undefined \def \showISBNxiii  #1{\unskip}     \fi
\ifx \showISSN     \undefined \def \showISSN      #1{\unskip}     \fi
\ifx \showLCCN     \undefined \def \showLCCN      #1{\unskip}     \fi
\ifx \shownote     \undefined \def \shownote      #1{#1}          \fi
\ifx \showarticletitle \undefined \def \showarticletitle #1{#1}   \fi
\ifx \showURL      \undefined \def \showURL       {\relax}        \fi
\providecommand\bibfield[2]{#2}
\providecommand\bibinfo[2]{#2}
\providecommand\natexlab[1]{#1}
\providecommand\showeprint[2][]{arXiv:#2}
\makeatletter
\@ifundefined{NAT@parse@date}{}{\let\NAT@parse@date@orig\NAT@parse@date}
\@ifundefined{NAT@parse@date}{}{\def\NAT@parse@date#1#2#3#4#5#6@@{\NAT@parse@date@orig#1#2#3#4#5#6@@\def\NAT@tempyear{0000}\def\NAT@tempexlab{{?}}\ifx\NAT@year\NAT@tempyear\ifx\NAT@exlab\NAT@tempexlab\def\NAT@date{[n.\,d.]}\else\edef\NAT@date{[n.\,d.]\NAT@exlab}\fi\fi}}
\makeatother

\bibitem[Asai et~al\mbox{.}(2023)]%
        {asai2023selfraglearningretrievegenerate}
\bibfield{author}{\bibinfo{person}{Akari Asai}, \bibinfo{person}{Zeqiu Wu}, \bibinfo{person}{Yizhong Wang}, \bibinfo{person}{Avirup Sil}, {and} \bibinfo{person}{Hannaneh Hajishirzi}.} \bibinfo{year}{2023}\natexlab{}.
\newblock \bibinfo{title}{Self-RAG: Learning to Retrieve, Generate, and Critique through Self-Reflection}.
\newblock
\showeprint[arxiv]{2310.11511}~[cs.CL]
\urldef\tempurl%
\url{https://arxiv.org/abs/2310.11511}
\showURL{%
\tempurl}


\bibitem[Babu and Indukuri(2026)]%
        {babu2026bindingdrift}
\bibfield{author}{\bibinfo{person}{Rahul~Suresh Babu} {and} \bibinfo{person}{Shashank Indukuri}.} \bibinfo{year}{2026}\natexlab{}.
\newblock \bibinfo{title}{Binding Drift in Multi-Step Tool-Augmented Agents}.
\newblock
\showeprint[arxiv]{2607.18316}~[cs.CL]
\urldef\tempurl%
\url{https://arxiv.org/abs/2607.18316}
\showURL{%
\tempurl}


\bibitem[Cao et~al\mbox{.}(2025)]%
        {cao2025multi2}
\bibfield{author}{\bibinfo{person}{Juntai Cao}, \bibinfo{person}{Xiang Zhang}, \bibinfo{person}{Raymond Li}, \bibinfo{person}{Jiaqi Wei}, \bibinfo{person}{Chuyuan Li}, \bibinfo{person}{Shafiq Joty}, {and} \bibinfo{person}{Giuseppe Carenini}.} \bibinfo{year}{2025}\natexlab{}.
\newblock \showarticletitle{Multi2: Multi-agent test-time scalable framework for multi-document processing}. In \bibinfo{booktitle}{\emph{Proceedings of The 5th New Frontiers in Summarization Workshop}}. \bibinfo{pages}{135--156}.
\newblock


\bibitem[Chen et~al\mbox{.}(2026)]%
        {chen2026you}
\bibfield{author}{\bibinfo{person}{Shengyuan Chen}, \bibinfo{person}{Chuang Zhou}, \bibinfo{person}{Zheng Yuan}, \bibinfo{person}{Qinggang Zhang}, \bibinfo{person}{Zeyang Cui}, \bibinfo{person}{Hao Chen}, \bibinfo{person}{Yilin Xiao}, \bibinfo{person}{Jiannong Cao}, {and} \bibinfo{person}{Xiao Huang}.} \bibinfo{year}{2026}\natexlab{}.
\newblock \showarticletitle{You don’t need pre-built graphs for rag: Retrieval augmented generation with adaptive reasoning structures}. In \bibinfo{booktitle}{\emph{Proceedings of the AAAI Conference on Artificial Intelligence}}, Vol.~\bibinfo{volume}{40}. \bibinfo{pages}{30270--30278}.
\newblock


\bibitem[Du et~al\mbox{.}(2026)]%
        {du2026rag}
\bibfield{author}{\bibinfo{person}{Mingxuan Du}, \bibinfo{person}{Benfeng Xu}, \bibinfo{person}{Chiwei Zhu}, \bibinfo{person}{Shaohan Wang}, \bibinfo{person}{Pengyu Wang}, \bibinfo{person}{Xiaorui Wang}, {and} \bibinfo{person}{Zhendong Mao}.} \bibinfo{year}{2026}\natexlab{}.
\newblock \showarticletitle{A-RAG: Scaling Agentic Retrieval-Augmented Generation via Hierarchical Retrieval Interfaces}.
\newblock \bibinfo{journal}{\emph{arXiv preprint arXiv:2602.03442}} (\bibinfo{year}{2026}).
\newblock


\bibitem[Edge et~al\mbox{.}(2024)]%
        {edge2024local}
\bibfield{author}{\bibinfo{person}{Darren Edge}, \bibinfo{person}{Ha Trinh}, \bibinfo{person}{Newman Cheng}, \bibinfo{person}{Joshua Bradley}, \bibinfo{person}{Alex Chao}, \bibinfo{person}{Apurva Mody}, \bibinfo{person}{Steven Truitt}, {and} \bibinfo{person}{Jonathan Larson}.} \bibinfo{year}{2024}\natexlab{}.
\newblock \showarticletitle{From local to global: A graph rag approach to query-focused summarization}.
\newblock \bibinfo{journal}{\emph{arXiv preprint arXiv:2404.16130}} (\bibinfo{year}{2024}).
\newblock


\bibitem[Gao et~al\mbox{.}(2023)]%
        {gao2023retrieval}
\bibfield{author}{\bibinfo{person}{Yunfan Gao}, \bibinfo{person}{Yun Xiong}, \bibinfo{person}{Xinyu Gao}, \bibinfo{person}{Kangxiang Jia}, \bibinfo{person}{Jinliu Pan}, \bibinfo{person}{Yuxi Bi}, \bibinfo{person}{Yi Dai}, \bibinfo{person}{Jiawei Sun}, {and} \bibinfo{person}{Haofen Wang}.} \bibinfo{year}{2023}\natexlab{}.
\newblock \showarticletitle{Retrieval-augmented generation for large language models: A survey}.
\newblock \bibinfo{journal}{\emph{arXiv preprint arXiv:2312.10997}} (\bibinfo{year}{2023}).
\newblock


\bibitem[Guo et~al\mbox{.}(2024)]%
        {guo2024lightrag}
\bibfield{author}{\bibinfo{person}{Zirui Guo}, \bibinfo{person}{Lianghao Xia}, \bibinfo{person}{Yanhua Yu}, \bibinfo{person}{Tu Ao}, {and} \bibinfo{person}{Chao Huang}.} \bibinfo{year}{2024}\natexlab{}.
\newblock \showarticletitle{LightRAG: Simple and Fast Retrieval-Augmented Generation}.
\newblock \bibinfo{journal}{\emph{arXiv preprint arXiv:2410.05779}} (\bibinfo{year}{2024}).
\newblock


\bibitem[Guo et~al\mbox{.}(2025)]%
        {lightrag}
\bibfield{author}{\bibinfo{person}{Zirui Guo}, \bibinfo{person}{Lianghao Xia}, \bibinfo{person}{Yanhua Yu}, \bibinfo{person}{Tu Ao}, {and} \bibinfo{person}{Chao Huang}.} \bibinfo{year}{2025}\natexlab{}.
\newblock \bibinfo{title}{LightRAG: Simple and Fast Retrieval-Augmented Generation}.
\newblock
\showeprint[arxiv]{2410.05779}~[cs.IR]
\urldef\tempurl%
\url{https://arxiv.org/abs/2410.05779}
\showURL{%
\tempurl}


\bibitem[Guti{\'e}rrez et~al\mbox{.}(2024)]%
        {hipporag}
\bibfield{author}{\bibinfo{person}{Bernal~Jim{\'e}nez Guti{\'e}rrez}, \bibinfo{person}{Yiheng Shu}, \bibinfo{person}{Yu Gu}, \bibinfo{person}{Michihiro Yasunaga}, {and} \bibinfo{person}{Yu Su}.} \bibinfo{year}{2024}\natexlab{}.
\newblock \showarticletitle{Hipporag: Neurobiologically inspired long-term memory for large language models}. In \bibinfo{booktitle}{\emph{The Thirty-eighth Annual Conference on Neural Information Processing Systems}}.
\newblock


\bibitem[Guti{\'e}rrez et~al\mbox{.}(2025a)]%
        {hippo2}
\bibfield{author}{\bibinfo{person}{Bernal~Jim{\'e}nez Guti{\'e}rrez}, \bibinfo{person}{Yiheng Shu}, \bibinfo{person}{Weijian Qi}, \bibinfo{person}{Sizhe Zhou}, {and} \bibinfo{person}{Yu Su}.} \bibinfo{year}{2025}\natexlab{a}.
\newblock \showarticletitle{From RAG to Memory: Non-Parametric Continual Learning for Large Language Models}. In \bibinfo{booktitle}{\emph{Forty-second International Conference on Machine Learning}}.
\newblock
\urldef\tempurl%
\url{https://openreview.net/forum?id=LWH8yn4HS2}
\showURL{%
\tempurl}


\bibitem[Guti{\'e}rrez et~al\mbox{.}(2025b)]%
        {hipporag2}
\bibfield{author}{\bibinfo{person}{Bernal~Jim{\'e}nez Guti{\'e}rrez}, \bibinfo{person}{Yiheng Shu}, \bibinfo{person}{Weijian Qi}, \bibinfo{person}{Sizhe Zhou}, {and} \bibinfo{person}{Yu Su}.} \bibinfo{year}{2025}\natexlab{b}.
\newblock \showarticletitle{From {RAG} to Memory: Non-Parametric Continual Learning for Large Language Models}. In \bibinfo{booktitle}{\emph{Forty-second International Conference on Machine Learning}}.
\newblock
\urldef\tempurl%
\url{https://openreview.net/forum?id=LWH8yn4HS2}
\showURL{%
\tempurl}


\bibitem[He et~al\mbox{.}(2024)]%
        {G-retriever}
\bibfield{author}{\bibinfo{person}{Xiaoxin He}, \bibinfo{person}{Yijun Tian}, \bibinfo{person}{Yifei Sun}, \bibinfo{person}{Nitesh Chawla}, \bibinfo{person}{Thomas Laurent}, \bibinfo{person}{Yann LeCun}, \bibinfo{person}{Xavier Bresson}, {and} \bibinfo{person}{Bryan Hooi}.} \bibinfo{year}{2024}\natexlab{}.
\newblock \showarticletitle{G-retriever: Retrieval-augmented generation for textual graph understanding and question answering}.
\newblock \bibinfo{journal}{\emph{Advances in Neural Information Processing Systems}}  \bibinfo{volume}{37} (\bibinfo{year}{2024}), \bibinfo{pages}{132876--132907}.
\newblock


\bibitem[Ho et~al\mbox{.}(2020)]%
        {2wiki}
\bibfield{author}{\bibinfo{person}{Xanh Ho}, \bibinfo{person}{Anh-Khoa Duong~Nguyen}, \bibinfo{person}{Saku Sugawara}, {and} \bibinfo{person}{Akiko Aizawa}.} \bibinfo{year}{2020}\natexlab{}.
\newblock \showarticletitle{Constructing A Multi-hop QA Dataset for Comprehensive Evaluation of Reasoning Steps}. In \bibinfo{booktitle}{\emph{Proceedings of the 28th International Conference on Computational Linguistics}}, \bibfield{editor}{\bibinfo{person}{Donia Scott}, \bibinfo{person}{Nuria Bel}, {and} \bibinfo{person}{Chengqing Zong}} (Eds.). \bibinfo{publisher}{International Committee on Computational Linguistics}, \bibinfo{address}{Barcelona, Spain (Online)}, \bibinfo{pages}{6609--6625}.
\newblock
\href{https://doi.org/10.18653/v1/2020.coling-main.580}{doi:\nolinkurl{10.18653/v1/2020.coling-main.580}}


\bibitem[Jiang et~al\mbox{.}(2023)]%
        {jiang2023active}
\bibfield{author}{\bibinfo{person}{Zhengbao Jiang}, \bibinfo{person}{Frank~F Xu}, \bibinfo{person}{Luyu Gao}, \bibinfo{person}{Zhiqing Sun}, \bibinfo{person}{Qian Liu}, \bibinfo{person}{Jane Dwivedi-Yu}, \bibinfo{person}{Yiming Yang}, \bibinfo{person}{Jamie Callan}, {and} \bibinfo{person}{Graham Neubig}.} \bibinfo{year}{2023}\natexlab{}.
\newblock \showarticletitle{Active retrieval augmented generation}. In \bibinfo{booktitle}{\emph{Proceedings of the 2023 Conference on Empirical Methods in Natural Language Processing}}.
\newblock


\bibitem[Jiao et~al\mbox{.}(2026)]%
        {jiao2026doctorrag}
\bibfield{author}{\bibinfo{person}{Shuguang Jiao}, \bibinfo{person}{Chengkai Huang}, \bibinfo{person}{Shuhan Qi}, \bibinfo{person}{Xuan Wang}, \bibinfo{person}{Yifan Li}, \bibinfo{person}{Quanchi Weng}, \bibinfo{person}{Lingchuan Liu}, \bibinfo{person}{Xunliang Cai}, {and} \bibinfo{person}{Lina Yao}.} \bibinfo{year}{2026}\natexlab{}.
\newblock \bibinfo{title}{Doctor-RAG: A Failure-Aware Repair Framework for Agentic Retrieval-Augmented Generation}.
\newblock
\showeprint[arxiv]{2604.00865}~[cs.IR]
\urldef\tempurl%
\url{https://arxiv.org/abs/2604.00865}
\showURL{%
\tempurl}


\bibitem[Jimenez~Gutierrez et~al\mbox{.}(2024)]%
        {jimenez2024hipporag}
\bibfield{author}{\bibinfo{person}{Bernal Jimenez~Gutierrez}, \bibinfo{person}{Yiheng Shu}, \bibinfo{person}{Yu Gu}, \bibinfo{person}{Michihiro Yasunaga}, {and} \bibinfo{person}{Yu Su}.} \bibinfo{year}{2024}\natexlab{}.
\newblock \showarticletitle{Hipporag: Neurobiologically inspired long-term memory for large language models}.
\newblock \bibinfo{journal}{\emph{Advances in Neural Information Processing Systems}}  \bibinfo{volume}{37} (\bibinfo{year}{2024}), \bibinfo{pages}{59532--59569}.
\newblock


\bibitem[Joren et~al\mbox{.}(2024)]%
        {joren2024sufficient}
\bibfield{author}{\bibinfo{person}{Hailey Joren}, \bibinfo{person}{Jianyi Zhang}, \bibinfo{person}{Chun-Sung Ferng}, \bibinfo{person}{Da-Cheng Juan}, \bibinfo{person}{Ankur Taly}, {and} \bibinfo{person}{Cyrus Rashtchian}.} \bibinfo{year}{2024}\natexlab{}.
\newblock \showarticletitle{Sufficient context: A new lens on retrieval augmented generation systems}.
\newblock \bibinfo{journal}{\emph{arXiv preprint arXiv:2411.06037}} (\bibinfo{year}{2024}).
\newblock


\bibitem[Kortukov et~al\mbox{.}(2024)]%
        {kortukov2024contextmemory}
\bibfield{author}{\bibinfo{person}{Evgenii Kortukov}, \bibinfo{person}{Alexander Rubinstein}, \bibinfo{person}{Elisa Nguyen}, {and} \bibinfo{person}{Seong~Joon Oh}.} \bibinfo{year}{2024}\natexlab{}.
\newblock \showarticletitle{Studying large language model behaviors under context-memory conflicts with real documents}.
\newblock \bibinfo{journal}{\emph{arXiv preprint arXiv:2404.16032}} (\bibinfo{year}{2024}).
\newblock


\bibitem[Lewis et~al\mbox{.}(2020)]%
        {lewis2020retrieval}
\bibfield{author}{\bibinfo{person}{Patrick Lewis}, \bibinfo{person}{Ethan Perez}, \bibinfo{person}{Aleksandra Piktus}, \bibinfo{person}{Fabio Petroni}, \bibinfo{person}{Vladimir Karpukhin}, \bibinfo{person}{Naman Goyal}, \bibinfo{person}{Heinrich K\"{u}ttler}, \bibinfo{person}{Mike Lewis}, \bibinfo{person}{Wen-tau Yih}, \bibinfo{person}{Tim Rockt\"{a}schel}, \bibinfo{person}{Sebastian Riedel}, {and} \bibinfo{person}{Douwe Kiela}.} \bibinfo{year}{2020}\natexlab{}.
\newblock \showarticletitle{Retrieval-augmented generation for knowledge-intensive NLP tasks}. In \bibinfo{booktitle}{\emph{Proceedings of the 34th International Conference on Neural Information Processing Systems}} (Vancouver, BC, Canada) \emph{(\bibinfo{series}{NIPS '20})}. \bibinfo{publisher}{Curran Associates Inc.}, \bibinfo{address}{Red Hook, NY, USA}, Article \bibinfo{articleno}{793}, \bibinfo{numpages}{16}~pages.
\newblock
\showISBNx{9781713829546}


\bibitem[Li et~al\mbox{.}(2026e)]%
        {li2026s2grag}
\bibfield{author}{\bibinfo{person}{Minghan Li}, \bibinfo{person}{Junjie Zou}, \bibinfo{person}{Xinxuan Lv}, \bibinfo{person}{Chao Zhang}, {and} \bibinfo{person}{Guodong Zhou}.} \bibinfo{year}{2026}\natexlab{e}.
\newblock \showarticletitle{{S}2{G}-{RAG}: Structured Sufficiency and Gap Judging for Iterative Retrieval-Augmented {QA}}. In \bibinfo{booktitle}{\emph{Proceedings of the 64th Annual Meeting of the {A}ssociation for {C}omputational {L}inguistics (Volume 1: Long Papers)}}, \bibfield{editor}{\bibinfo{person}{Maria Liakata}, \bibinfo{person}{Viviane~P. Moreira}, \bibinfo{person}{Jiajun Zhang}, {and} \bibinfo{person}{David Jurgens}} (Eds.). \bibinfo{publisher}{Association for Computational Linguistics}, \bibinfo{address}{San Diego, California, United States}, \bibinfo{pages}{25846--25862}.
\newblock
\showISBNx{979-8-89176-390-6}
\href{https://doi.org/10.18653/v1/2026.acl-long.1185}{doi:\nolinkurl{10.18653/v1/2026.acl-long.1185}}


\bibitem[Li et~al\mbox{.}(2026a)]%
        {li2026draftrl}
\bibfield{author}{\bibinfo{person}{Yuanhao Li}, \bibinfo{person}{Mingshan Liu}, \bibinfo{person}{Hongbo Wang}, \bibinfo{person}{Yiding Zhang}, \bibinfo{person}{Yifei Ma}, {and} \bibinfo{person}{Wei Tan}.} \bibinfo{year}{2026}\natexlab{a}.
\newblock \showarticletitle{{DRAFT-RL: Multi-Agent Chain-of-Draft Reasoning for Reinforcement Learning-Enhanced LLMs}}.
\newblock \bibinfo{journal}{\emph{Proceedings of the AAAI Conference on Artificial Intelligence}} \bibinfo{volume}{40}, \bibinfo{number}{35} (\bibinfo{year}{2026}), \bibinfo{pages}{29530--29537}.
\newblock
\href{https://doi.org/10.1609/aaai.v40i35.40195}{doi:\nolinkurl{10.1609/aaai.v40i35.40195}}


\bibitem[Li et~al\mbox{.}(2026b)]%
        {li2026retrievalaugmentedfakenews}
\bibfield{author}{\bibinfo{person}{Yuanhao Li}, \bibinfo{person}{Hongbo Wang}, \bibinfo{person}{Xuhong Chen}, {and} \bibinfo{person}{Yiming Cao}.} \bibinfo{year}{2026}\natexlab{b}.
\newblock \showarticletitle{{Retrieval-Augmented Multi-Agent Multimodal Framework for Fake News Detection}}. In \bibinfo{booktitle}{\emph{ICASSP 2026}}.
\newblock
\urldef\tempurl%
\url{https://cmsworkshops.com/ICASSP2026/view_paper.php?PaperNum=9709}
\showURL{%
\tempurl}


\bibitem[Li et~al\mbox{.}(2026c)]%
        {li2026counterfactualrolloutreplayforkable}
\bibfield{author}{\bibinfo{person}{Yuanhao Li}, \bibinfo{person}{Hongbo Wang}, \bibinfo{person}{Xuhong Chen}, \bibinfo{person}{Yiming Cao}, {and} \bibinfo{person}{Xunzhu Tang}.} \bibinfo{year}{2026}\natexlab{c}.
\newblock \bibinfo{title}{{Counterfactual Rollout Replay: Forkable Environments as Free Process Rewards for Software Engineering Agents}}.
\newblock
\showeprint[arxiv]{2609.33875}~[cs.SE]
\urldef\tempurl%
\url{https://arxiv.org/abs/2609.33875}
\showURL{%
\tempurl}
\newblock
\shownote{Accepted at NeurIPS 2026}.


\bibitem[Li et~al\mbox{.}(2026d)]%
        {li2026boostaprboostingautomatedprogram}
\bibfield{author}{\bibinfo{person}{Yuanhao Li}, \bibinfo{person}{Hongbo Wang}, \bibinfo{person}{Xiaotang Shang}, \bibinfo{person}{Xunzhu Tang}, \bibinfo{person}{Yiming Cao}, {and} \bibinfo{person}{Xuhong Chen}.} \bibinfo{year}{2026}\natexlab{d}.
\newblock \bibinfo{title}{{BoostAPR: Boosting Automated Program Repair via Execution-Grounded Reinforcement Learning with Dual Reward Models}}.
\newblock
\showeprint[arxiv]{2605.09134}~[cs.AI]
\urldef\tempurl%
\url{https://arxiv.org/abs/2605.09134}
\showURL{%
\tempurl}
\newblock
\shownote{Accepted at ICML 2026}.


\bibitem[Li et~al\mbox{.}(2025)]%
        {li2024structrag}
\bibfield{author}{\bibinfo{person}{Zhuoqun Li}, \bibinfo{person}{Xuanang Chen}, \bibinfo{person}{Haiyang Yu}, \bibinfo{person}{Hongyu Lin}, \bibinfo{person}{Yaojie Lu}, \bibinfo{person}{Qiaoyu Tang}, \bibinfo{person}{Fei Huang}, \bibinfo{person}{Xianpei Han}, \bibinfo{person}{Le Sun}, {and} \bibinfo{person}{Yongbin Li}.} \bibinfo{year}{2025}\natexlab{}.
\newblock \showarticletitle{Struct{RAG}: Boosting Knowledge Intensive Reasoning of {LLM}s via Inference-time Hybrid Information Structurization}. In \bibinfo{booktitle}{\emph{The Thirteenth International Conference on Learning Representations}}.
\newblock
\urldef\tempurl%
\url{https://openreview.net/forum?id=GhexuBLxbO}
\showURL{%
\tempurl}


\bibitem[Luo et~al\mbox{.}(2025)]%
        {GFM}
\bibfield{author}{\bibinfo{person}{Linhao Luo}, \bibinfo{person}{Zicheng Zhao}, \bibinfo{person}{Gholamreza Haffari}, \bibinfo{person}{Dinh Phung}, \bibinfo{person}{Chen Gong}, {and} \bibinfo{person}{Shirui Pan}.} \bibinfo{year}{2025}\natexlab{}.
\newblock \showarticletitle{GFM-RAG: Graph Foundation Model for Retrieval Augmented Generation}.
\newblock \bibinfo{journal}{\emph{arXiv preprint arXiv:2502.01113}} (\bibinfo{year}{2025}).
\newblock


\bibitem[OpenAI(2024)]%
        {gpt}
\bibfield{author}{\bibinfo{person}{OpenAI}.} \bibinfo{year}{2024}\natexlab{}.
\newblock \bibinfo{title}{GPT-4o System Card}.
\newblock
\showeprint[arxiv]{2410.21276}~[cs.CL]
\urldef\tempurl%
\url{https://arxiv.org/abs/2410.21276}
\showURL{%
\tempurl}


\bibitem[Qiu et~al\mbox{.}(2026)]%
        {qiu2026surerag}
\bibfield{author}{\bibinfo{person}{Jingxi Qiu}, \bibinfo{person}{Zeyu Han}, {and} \bibinfo{person}{Cheng Huang}.} \bibinfo{year}{2026}\natexlab{}.
\newblock \bibinfo{title}{SURE-RAG: Sufficiency and Uncertainty-Aware Evidence Verification for Selective Retrieval-Augmented Generation}.
\newblock
\showeprint[arxiv]{2605.03534}~[cs.CL]
\urldef\tempurl%
\url{https://arxiv.org/abs/2605.03534}
\showURL{%
\tempurl}


\bibitem[Ru et~al\mbox{.}(2024)]%
        {ru2024ragchecker}
\bibfield{author}{\bibinfo{person}{Dongyu Ru}, \bibinfo{person}{Lin Qiu}, \bibinfo{person}{Xiangkun Hu}, \bibinfo{person}{Tianhang Zhang}, \bibinfo{person}{Peng Shi}, \bibinfo{person}{Shuaichen Chang}, \bibinfo{person}{Cheng Jiayang}, \bibinfo{person}{Cunxiang Wang}, \bibinfo{person}{Shichao Sun}, \bibinfo{person}{Huanyu Li}, {et~al\mbox{.}}} \bibinfo{year}{2024}\natexlab{}.
\newblock \showarticletitle{Ragchecker: A fine-grained framework for diagnosing retrieval-augmented generation}.
\newblock \bibinfo{journal}{\emph{Advances in Neural Information Processing Systems}}  \bibinfo{volume}{37} (\bibinfo{year}{2024}), \bibinfo{pages}{21999--22027}.
\newblock


\bibitem[Sarthi et~al\mbox{.}(2024a)]%
        {sarthi2024raptor}
\bibfield{author}{\bibinfo{person}{Parth Sarthi}, \bibinfo{person}{Salman Abdullah}, \bibinfo{person}{Aditi Tuli}, \bibinfo{person}{Shubh Khanna}, \bibinfo{person}{Anna Goldie}, {and} \bibinfo{person}{Christopher~D. Manning}.} \bibinfo{year}{2024}\natexlab{a}.
\newblock \showarticletitle{RAPTOR: Recursive Abstractive Processing for Tree-Organized Retrieval}. In \bibinfo{booktitle}{\emph{International Conference on Learning Representations (ICLR)}}.
\newblock


\bibitem[Sarthi et~al\mbox{.}(2024b)]%
        {raptor}
\bibfield{author}{\bibinfo{person}{Parth Sarthi}, \bibinfo{person}{Salman Abdullah}, \bibinfo{person}{Aditi Tuli}, \bibinfo{person}{Shubh Khanna}, \bibinfo{person}{Anna Goldie}, {and} \bibinfo{person}{Christopher~D Manning}.} \bibinfo{year}{2024}\natexlab{b}.
\newblock \showarticletitle{Raptor: Recursive abstractive processing for tree-organized retrieval}. In \bibinfo{booktitle}{\emph{The Twelfth International Conference on Learning Representations}}.
\newblock


\bibitem[Tan et~al\mbox{.}(2026)]%
        {tan2026evocurl}
\bibfield{author}{\bibinfo{person}{Wei Tan}, \bibinfo{person}{Yuanhao Li}, {and} \bibinfo{person}{Wenkai Liang}.} \bibinfo{year}{2026}\natexlab{}.
\newblock \showarticletitle{{Evo-CuRL: Curriculum-Aware Reinforcement Learning over Code Lineage Graphs for Software Engineering Reasoning}}. In \bibinfo{booktitle}{\emph{ICMR 2026}}. \bibinfo{pages}{1327--1335}.
\newblock
\href{https://doi.org/10.1145/3805622.3810701}{doi:\nolinkurl{10.1145/3805622.3810701}}


\bibitem[Trivedi et~al\mbox{.}(2022)]%
        {musi}
\bibfield{author}{\bibinfo{person}{Harsh Trivedi}, \bibinfo{person}{Niranjan Balasubramanian}, \bibinfo{person}{Tushar Khot}, {and} \bibinfo{person}{Ashish Sabharwal}.} \bibinfo{year}{2022}\natexlab{}.
\newblock \showarticletitle{MuSiQue: Multihop Questions via Single-hop Question Composition}.
\newblock \bibinfo{journal}{\emph{Trans. Assoc. Comput. Linguistics}}  \bibinfo{volume}{10} (\bibinfo{year}{2022}), \bibinfo{pages}{539--554}.
\newblock
\href{https://doi.org/10.1162/TACL\_A\_00475}{doi:\nolinkurl{10.1162/TACL\_A\_00475}}


\bibitem[Trivedi et~al\mbox{.}(2023)]%
        {IRCOT}
\bibfield{author}{\bibinfo{person}{Harsh Trivedi}, \bibinfo{person}{Niranjan Balasubramanian}, \bibinfo{person}{Tushar Khot}, {and} \bibinfo{person}{Ashish Sabharwal}.} \bibinfo{year}{2023}\natexlab{}.
\newblock \showarticletitle{Interleaving Retrieval with Chain-of-Thought Reasoning for Knowledge-Intensive Multi-Step Questions}. In \bibinfo{booktitle}{\emph{The 61st Annual Meeting Of The Association For Computational Linguistics}}.
\newblock


\bibitem[Wang et~al\mbox{.}(2024)]%
        {KGP}
\bibfield{author}{\bibinfo{person}{Yu Wang}, \bibinfo{person}{Nedim Lipka}, \bibinfo{person}{Ryan~A Rossi}, \bibinfo{person}{Alexa Siu}, \bibinfo{person}{Ruiyi Zhang}, {and} \bibinfo{person}{Tyler Derr}.} \bibinfo{year}{2024}\natexlab{}.
\newblock \showarticletitle{Knowledge graph prompting for multi-document question answering}. In \bibinfo{booktitle}{\emph{Proceedings of the AAAI Conference on Artificial Intelligence}}, Vol.~\bibinfo{volume}{38}. \bibinfo{pages}{19206--19214}.
\newblock


\bibitem[Wei et~al\mbox{.}(2026a)]%
        {wei2026think}
\bibfield{author}{\bibinfo{person}{Jiaqi Wei}, \bibinfo{person}{Xuehang Guo}, \bibinfo{person}{Pengfei Yu}, \bibinfo{person}{Xiang Zhang}, \bibinfo{person}{Wanli Ouyang}, \bibinfo{person}{Siqi Sun}, \bibinfo{person}{Qingyun Wang}, {and} \bibinfo{person}{Chenyu You}.} \bibinfo{year}{2026}\natexlab{a}.
\newblock \showarticletitle{When to Think, When to Speak: Learning Disclosure Policies for LLM Reasoning}.
\newblock \bibinfo{journal}{\emph{arXiv preprint arXiv:2605.03314}} (\bibinfo{year}{2026}).
\newblock


\bibitem[Wei et~al\mbox{.}(2022)]%
        {cot}
\bibfield{author}{\bibinfo{person}{Jason Wei}, \bibinfo{person}{Xuezhi Wang}, \bibinfo{person}{Dale Schuurmans}, \bibinfo{person}{Maarten Bosma}, \bibinfo{person}{brian ichter}, \bibinfo{person}{Fei Xia}, \bibinfo{person}{Ed Chi}, \bibinfo{person}{Quoc~V Le}, {and} \bibinfo{person}{Denny Zhou}.} \bibinfo{year}{2022}\natexlab{}.
\newblock \showarticletitle{Chain-of-Thought Prompting Elicits Reasoning in Large Language Models}. In \bibinfo{booktitle}{\emph{Advances in Neural Information Processing Systems}}, \bibfield{editor}{\bibinfo{person}{S.~Koyejo}, \bibinfo{person}{S.~Mohamed}, \bibinfo{person}{A.~Agarwal}, \bibinfo{person}{D.~Belgrave}, \bibinfo{person}{K.~Cho}, {and} \bibinfo{person}{A.~Oh}} (Eds.), Vol.~\bibinfo{volume}{35}. \bibinfo{publisher}{Curran Associates, Inc.}, \bibinfo{pages}{24824--24837}.
\newblock
\urldef\tempurl%
\url{https://proceedings.neurips.cc/paper_files/paper/2022/file/9d5609613524ecf4f15af0f7b31abca4-Paper-Conference.pdf}
\showURL{%
\tempurl}


\bibitem[Wei et~al\mbox{.}(2025a)]%
        {wei2025ai}
\bibfield{author}{\bibinfo{person}{Jiaqi Wei}, \bibinfo{person}{Yuejin Yang}, \bibinfo{person}{Xiang Zhang}, \bibinfo{person}{Yuhan Chen}, \bibinfo{person}{Xiang Zhuang}, \bibinfo{person}{Zhangyang Gao}, \bibinfo{person}{Dongzhan Zhou}, \bibinfo{person}{Guangshuai Wang}, \bibinfo{person}{Zhiqiang Gao}, \bibinfo{person}{Juntai Cao}, {et~al\mbox{.}}} \bibinfo{year}{2025}\natexlab{a}.
\newblock \showarticletitle{From ai for science to agentic science: A survey on autonomous scientific discovery}.
\newblock \bibinfo{journal}{\emph{arXiv preprint arXiv:2508.14111}} (\bibinfo{year}{2025}).
\newblock


\bibitem[Wei et~al\mbox{.}(2026c)]%
        {wei2026llm}
\bibfield{author}{\bibinfo{person}{Jiaqi Wei}, \bibinfo{person}{Xiang Zhang}, \bibinfo{person}{Yuejin Yang}, \bibinfo{person}{Wenxuan Huang}, \bibinfo{person}{Juntai Cao}, \bibinfo{person}{Sheng Xu}, \bibinfo{person}{Xiang Zhuang}, \bibinfo{person}{Zhangyang Gao}, \bibinfo{person}{Muhammad Abdul-Mageed}, \bibinfo{person}{Laks~VS Lakshmanan}, {et~al\mbox{.}}} \bibinfo{year}{2026}\natexlab{c}.
\newblock \showarticletitle{When LLM Meets Tree Search: A Systematic View of Inference as Search in Large Language Models}.
\newblock \bibinfo{journal}{\emph{arXiv preprint arXiv:2608.30395}} (\bibinfo{year}{2026}).
\newblock


\bibitem[Wei et~al\mbox{.}(2025b)]%
        {weiretrieval}
\bibfield{author}{\bibinfo{person}{Jiaqi Wei}, \bibinfo{person}{Hao Zhou}, \bibinfo{person}{Xiang Zhang}, \bibinfo{person}{Di Zhang}, \bibinfo{person}{Zijie Qiu}, \bibinfo{person}{Noah Wei}, \bibinfo{person}{Jinzhe Li}, \bibinfo{person}{Wanli Ouyang}, {and} \bibinfo{person}{Siqi Sun}.} \bibinfo{year}{2025}\natexlab{b}.
\newblock \showarticletitle{Retrieval is Not Enough: Enhancing RAG through Test-Time Critique and Optimization}. In \bibinfo{booktitle}{\emph{The Thirty-ninth Annual Conference on Neural Information Processing Systems}}.
\newblock


\bibitem[Wei et~al\mbox{.}(2026b)]%
        {wei2026skillrag}
\bibfield{author}{\bibinfo{person}{Kai Wei}, \bibinfo{person}{Raymond Li}, \bibinfo{person}{Xi Zhu}, \bibinfo{person}{Zhaoqian Xue}, \bibinfo{person}{Jiaojiao Han}, \bibinfo{person}{Jingcheng Niu}, {and} \bibinfo{person}{Fan Yang}.} \bibinfo{year}{2026}\natexlab{b}.
\newblock \bibinfo{title}{Skill-RAG: Failure-State-Aware Retrieval Augmentation via Hidden-State Probing and Skill Routing}.
\newblock
\showeprint[arxiv]{2604.15771}~[cs.CL]
\urldef\tempurl%
\url{https://arxiv.org/abs/2604.15771}
\showURL{%
\tempurl}


\bibitem[Xiao et~al\mbox{.}(2025)]%
        {xiao2025lag}
\bibfield{author}{\bibinfo{person}{Yilin Xiao}, \bibinfo{person}{Chuang Zhou}, \bibinfo{person}{Yujing Zhang}, \bibinfo{person}{Qinggang Zhang}, \bibinfo{person}{Su Dong}, \bibinfo{person}{Shengyuan Chen}, \bibinfo{person}{Chang Yang}, {and} \bibinfo{person}{Xiao Huang}.} \bibinfo{year}{2025}\natexlab{}.
\newblock \showarticletitle{Lag: Logic-augmented generation from a cartesian perspective}.
\newblock \bibinfo{journal}{\emph{arXiv preprint arXiv:2508.05509}} (\bibinfo{year}{2025}).
\newblock


\bibitem[Xu et~al\mbox{.}(2026a)]%
        {xu2026evomas}
\bibfield{author}{\bibinfo{person}{Chengdong Xu}, \bibinfo{person}{Kaiqiang Ke}, \bibinfo{person}{Ziheng Liu}, \bibinfo{person}{Jiaqi Wei}, \bibinfo{person}{Zibo Shao}, \bibinfo{person}{Weile Guo}, {and} \bibinfo{person}{Chao Yu}.} \bibinfo{year}{2026}\natexlab{a}.
\newblock \showarticletitle{EvoMAS: Learning Execution-Time Workflows for Multi-Agent Systems}.
\newblock \bibinfo{journal}{\emph{arXiv preprint arXiv:2605.08769}} (\bibinfo{year}{2026}).
\newblock


\bibitem[Xu et~al\mbox{.}(2026b)]%
        {xu2026agent}
\bibfield{author}{\bibinfo{person}{Nuoyue Xu}, \bibinfo{person}{Jiang Liu}, \bibinfo{person}{Wenxuan Huang}, \bibinfo{person}{Xiang Zhang}, \bibinfo{person}{Juntai Cao}, {and} \bibinfo{person}{Jiaqi Wei}.} \bibinfo{year}{2026}\natexlab{b}.
\newblock \showarticletitle{Agent Memory with Episodic Retrieval for Financial Decision-Making}.
\newblock \bibinfo{journal}{\emph{arXiv preprint arXiv:2609.28771}} (\bibinfo{year}{2026}).
\newblock


\bibitem[Yan et~al\mbox{.}(2024)]%
        {yan2024corrective}
\bibfield{author}{\bibinfo{person}{Shi-Qi Yan}, \bibinfo{person}{Jia-Chen Gu}, \bibinfo{person}{Yun Zhu}, {and} \bibinfo{person}{Zhen-Hua Ling}.} \bibinfo{year}{2024}\natexlab{}.
\newblock \bibinfo{title}{Corrective Retrieval Augmented Generation}.
\newblock
\urldef\tempurl%
\url{https://openreview.net/forum?id=JnWJbrnaUE}
\showURL{%
\tempurl}


\bibitem[Yang et~al\mbox{.}(2018)]%
        {hotpotqa}
\bibfield{author}{\bibinfo{person}{Zhilin Yang}, \bibinfo{person}{Peng Qi}, \bibinfo{person}{Saizheng Zhang}, \bibinfo{person}{Yoshua Bengio}, \bibinfo{person}{William Cohen}, \bibinfo{person}{Ruslan Salakhutdinov}, {and} \bibinfo{person}{Christopher~D. Manning}.} \bibinfo{year}{2018}\natexlab{}.
\newblock \showarticletitle{HotpotQA: A Dataset for Diverse, Explainable Multi-hop Question Answering}. In \bibinfo{booktitle}{\emph{Proceedings of the 2018 Conference on Empirical Methods in Natural Language Processing}}, \bibfield{editor}{\bibinfo{person}{Ellen Riloff}, \bibinfo{person}{David Chiang}, \bibinfo{person}{Julia Hockenmaier}, {and} \bibinfo{person}{Jun{'}ichi Tsujii}} (Eds.). \bibinfo{publisher}{Association for Computational Linguistics}, \bibinfo{address}{Brussels, Belgium}, \bibinfo{pages}{2369--2380}.
\newblock
\href{https://doi.org/10.18653/v1/D18-1259}{doi:\nolinkurl{10.18653/v1/D18-1259}}


\bibitem[Yang et~al\mbox{.}(2026)]%
        {yang2026forestllm}
\bibfield{author}{\bibinfo{person}{Zhihan Yang}, \bibinfo{person}{Jiaqi Wei}, \bibinfo{person}{Xiang Zhang}, \bibinfo{person}{Haoyu Dong}, \bibinfo{person}{Yiwen Wang}, \bibinfo{person}{Xiaoke Guo}, \bibinfo{person}{Pengkun Zhang}, \bibinfo{person}{Yiwei Xu}, {and} \bibinfo{person}{Chenyu You}.} \bibinfo{year}{2026}\natexlab{}.
\newblock \showarticletitle{FORESTLLM: Large Language Models Make Random Forest Great on Few-shot Tabular Learning}.
\newblock \bibinfo{journal}{\emph{arXiv preprint arXiv:2601.11311}} (\bibinfo{year}{2026}).
\newblock


\bibitem[Yu et~al\mbox{.}(2024)]%
        {yu2023chainofnote}
\bibfield{author}{\bibinfo{person}{Wenhao Yu}, \bibinfo{person}{Hongming Zhang}, \bibinfo{person}{Xiaoman Pan}, \bibinfo{person}{Peixin Cao}, \bibinfo{person}{Kaixin Ma}, \bibinfo{person}{Jian Li}, \bibinfo{person}{Hongwei Wang}, {and} \bibinfo{person}{Dong Yu}.} \bibinfo{year}{2024}\natexlab{}.
\newblock \showarticletitle{Chain-of-Note: Enhancing Robustness in Retrieval-Augmented Language Models}. In \bibinfo{booktitle}{\emph{Proceedings of the 2024 Conference on Empirical Methods in Natural Language Processing}}, \bibfield{editor}{\bibinfo{person}{Yaser Al-Onaizan}, \bibinfo{person}{Mohit Bansal}, {and} \bibinfo{person}{Yun-Nung Chen}} (Eds.). \bibinfo{publisher}{Association for Computational Linguistics}, \bibinfo{address}{Miami, Florida, USA}, \bibinfo{pages}{14672--14685}.
\newblock
\href{https://doi.org/10.18653/v1/2024.emnlp-main.813}{doi:\nolinkurl{10.18653/v1/2024.emnlp-main.813}}


\bibitem[Zhang et~al\mbox{.}(2025a)]%
        {zhang2025tokenization}
\bibfield{author}{\bibinfo{person}{Xiang Zhang}, \bibinfo{person}{Juntai Cao}, \bibinfo{person}{Jiaqi Wei}, \bibinfo{person}{Yiwei Xu}, {and} \bibinfo{person}{Chenyu You}.} \bibinfo{year}{2025}\natexlab{a}.
\newblock \showarticletitle{Tokenization constraints in llms: A study of symbolic and arithmetic reasoning limits}.
\newblock \bibinfo{journal}{\emph{arXiv preprint arXiv:2505.14178}} (\bibinfo{year}{2025}).
\newblock


\bibitem[Zhang et~al\mbox{.}(2025b)]%
        {zhang2025does}
\bibfield{author}{\bibinfo{person}{Xiang Zhang}, \bibinfo{person}{Juntai Cao}, \bibinfo{person}{Jiaqi Wei}, \bibinfo{person}{Chenyu You}, {and} \bibinfo{person}{Dujian Ding}.} \bibinfo{year}{2025}\natexlab{b}.
\newblock \showarticletitle{Why does your cot prompt (not) work? theoretical analysis of prompt space complexity, its interaction with answer space during cot reasoning with llms: A recurrent perspective}.
\newblock \bibinfo{journal}{\emph{arXiv preprint arXiv:2503.10084}} (\bibinfo{year}{2025}).
\newblock


\bibitem[Zhang et~al\mbox{.}(2025c)]%
        {zhang2025postergen}
\bibfield{author}{\bibinfo{person}{Zhilin Zhang}, \bibinfo{person}{Xiang Zhang}, \bibinfo{person}{Jiaqi Wei}, \bibinfo{person}{Yiwei Xu}, {and} \bibinfo{person}{Chenyu You}.} \bibinfo{year}{2025}\natexlab{c}.
\newblock \showarticletitle{Postergen: Aesthetic-aware paper-to-poster generation via multi-agent llms}.
\newblock \bibinfo{journal}{\emph{arXiv e-prints}} (\bibinfo{year}{2025}), \bibinfo{pages}{arXiv--2508}.
\newblock


\bibitem[Zhou et~al\mbox{.}(2025)]%
        {zhou2025scientists}
\bibfield{author}{\bibinfo{person}{Yuhao Zhou}, \bibinfo{person}{Yiheng Wang}, \bibinfo{person}{Xuming He}, \bibinfo{person}{Ao Shen}, \bibinfo{person}{Ruoyao Xiao}, \bibinfo{person}{Zhiwei Li}, \bibinfo{person}{Qiantai Feng}, \bibinfo{person}{Zijie Guo}, \bibinfo{person}{Yuejin Yang}, \bibinfo{person}{Hao Wu}, {et~al\mbox{.}}} \bibinfo{year}{2025}\natexlab{}.
\newblock \showarticletitle{Scientists' First Exam: Probing Cognitive Abilities of MLLM via Perception, Understanding, and Reasoning}.
\newblock \bibinfo{journal}{\emph{arXiv preprint arXiv:2506.10521}} (\bibinfo{year}{2025}).
\newblock


\bibitem[Zhuang et~al\mbox{.}(2025)]%
        {linear}
\bibfield{author}{\bibinfo{person}{Luyao Zhuang}, \bibinfo{person}{Shengyuan Chen}, \bibinfo{person}{Yilin Xiao}, \bibinfo{person}{Huachi Zhou}, \bibinfo{person}{Yujing Zhang}, \bibinfo{person}{Hao Chen}, \bibinfo{person}{Qinggang Zhang}, {and} \bibinfo{person}{Xiao Huang}.} \bibinfo{year}{2025}\natexlab{}.
\newblock \bibinfo{title}{LinearRAG: Linear Graph Retrieval Augmented Generation on Large-scale Corpora}.
\newblock
\showeprint[arxiv]{2510.10114}~[cs.CL]
\urldef\tempurl%
\url{https://arxiv.org/abs/2510.10114}
\showURL{%
\tempurl}


\end{thebibliography}

\clearpage
\appendix

\section*{Appendix Overview}
\label{app:overview}

This appendix is organized as follows:
\begin{itemize}[leftmargin=1.6em, itemsep=0.3em, topsep=0.3em]

    \item \hyperref[app:alg]{\textbf{Appendix A: Algorithm}.}
    Full inference procedure of {\methodname}, including register construction, register updates, progress-delta injection, and budget-aware control.

    \item \hyperref[app:impl]{\textbf{Appendix B: Extended Implementation Details}.}
    Additional details on model configuration, retrieval tools, budgets, deduplication, progress-delta injection, and other engineering choices.

    \item \hyperref[app:category]{\textbf{Appendix C: Per-Category Breakdown}.}
    Category-level results for HotpotQA, 2WikiMultiHopQA, and MuSiQue under both embedding backbones.

    \item \hyperref[app:efficiency]{\textbf{Appendix D: Retrieval Efficiency Analysis}.}
    Retrieved tokens and API cost across methods, ablations, and embedding backbones.

    \item \hyperref[app:error]{\textbf{Appendix E: Failure Case Analysis}.}
    Error breakdown by type, hop count, and embedding model.

    \item \hyperref[app:case]{\textbf{Appendix F: Case Study: How the Register Guides the Search}.}
    Qualitative examples showing how the register keeps the search on target and supports self-correction.

    \item \hyperref[app:prompts]{\textbf{Appendix G: Prompt Templates}.}
    Key prompts used in {\methodname}, including initialization, update, budget warning, and force-answer prompts.

    \item \hyperref[app:broader]{\textbf{Appendix H: Broader Impacts}.}
    Discussion of potential applications, transparency benefits, and responsible usage considerations.

\end{itemize}

\section{Algorithm}
\label{app:alg}

\begin{algorithm}[!ht]
\caption{{\methodname} Inference}
\label{alg:method_inference}
\begin{algorithmic}[1]
\Require Query $q$, corpus $\mathcal{K}$, retrieval budget $B$
\Ensure Final answer $a$

\State $\mathcal{C}^{(0)} \gets R_{\text{seed}}(q; \mathcal{K})$
\State $\mathcal{S}_0 \gets \textsc{InitializeState}(q, \mathcal{C}^{(0)})$
\State Inject the initial register summary of $\mathcal{S}_0$ into the agent context

\For{$t = 1$ to $B$}
    \If{$\textsc{ReadyToAnswer}(q, \mathcal{S}_{t-1})$}
        \State \Return $G(q, \mathcal{S}_{t-1})$
    \EndIf

    \State $b_t \gets \textsc{SelectActiveRequirement}(\mathcal{S}_{t-1})$
    \State $u_t \gets \textsc{Compose}(q, d_{b_t}, e_{b_t}^{(t-1)})$
    \State $\Delta \mathcal{C}_t \gets R(u_t; \mathcal{K})$

    \If{$\Delta \mathcal{C}_t = \varnothing$}
        \State $\mathcal{S}_t \gets \mathcal{S}_{t-1}$
    \Else
        \State $\tilde{\mathcal{S}}_t \gets U(\mathcal{S}_{t-1}, \Delta \mathcal{C}_t)$
        \State $\mathcal{S}_t \gets \textsc{ApplyInvariants}(\tilde{\mathcal{S}}_t, \mathcal{S}_{t-1})$
    \EndIf

    \If{\textsc{HasProgressDelta}($\mathcal{S}_{t-1}, \mathcal{S}_t$)}
        \State Inject progress delta into the agent context
    \EndIf

    \If{$t \ge 0.8B$}
        \State Inject budget-aware nudge
    \EndIf
\EndFor

\State \Return $G(q, \mathcal{S}_B)$
\end{algorithmic}
\end{algorithm}

\section{Extended Implementation Details}
\label{app:impl}

\paragraph{Language model}
All experiments use GPT-4o-mini as the default LLM.
Structured JSON calls for seed analysis and register updates use
\texttt{temperature\,=\,0.0} to reduce output variance, while the tool-using agent loop uses the model's default sampling settings.
The maximum output length for each LLM call is 16{,}384 tokens.

\paragraph{Retrieval tools}
The agent has access to three tools followed by A-RAG~\cite{du2026rag}:
\texttt{semantic\_search} for dense cosine retrieval with the active embedding model
(default top-$k=5$),
\texttt{keyword\_search} for BM25-style lexical retrieval
(default top-$k=5$),
and \texttt{read\_chunk} for direct chunk lookup by identifier.
Seed retrieval uses \texttt{semantic\_search} with top-$k=5$.
Retrieved chunks are deduplicated within each loop, and previously read chunks are not re-exposed to the register update prompt.

\paragraph{Embedding models}
The primary embedding model is \textbf{all-MiniLM-L6-v2}, a 384-dimensional bi-encoder optimized for semantic similarity.
The alternative backbone is \textbf{Qwen3-Embedding-0.6B}, a 1024-dimensional embedding model.
Embeddings are pre-computed and stored as flat similarity indices; no re-indexing is performed between runs.

\paragraph{Evaluation corpus}
We follow the HippoRAG evaluation protocol, using the same retrieval corpora and the same 1{,}000-question splits sampled from the official validation sets of HotpotQA, 2WikiMultiHopQA, and MuSiQue.
All baselines are evaluated on identical question sets.

\paragraph{Retrieval budget}
The agent is allowed at most 15 retrieval loops per question.
We also enforce a cumulative context budget of 128{,}000 tokens.
If this budget is exceeded, a force-answer prompt is injected and further tool calls are disabled, yielding a budget-exhausted answer, categorized as ``Gave Up'' in the error analysis.

\paragraph{Search deduplication}
Each retrieval action is cached by a normalized key: sorted keywords for \texttt{keyword\_search} and the lowercased query string for \texttt{semantic\_search}.
If the agent issues an identical query, the system returns a deduplication notice with the previously retrieved chunk IDs, prompting the agent to read existing chunks or reformulate the query.
This reduces cyclic search and unnecessary token usage.

\paragraph{Progress delta injection}
Rather than replaying the full register after every step, the system injects a compact \texttt{[Progress]} delta only when at least one item changes materially.
The delta records the changed items with abbreviated evidence notes and a one-line summary of the items still open.
On 4-hop questions, this reduces register-injection overhead by approximately 40\% relative to replaying the full list.

\paragraph{Budget-aware nudge}
When the agent reaches 80\% of its loop budget, a soft warning is injected to encourage synthesis if the current evidence is sufficient, or to make the remaining retrieval steps more targeted.
This reduces redundant last-mile searches near budget exhaustion.

\section{Per-Category Breakdown}
\label{app:category}

Tables~\ref{tab:musique_minilm}--\ref{tab:2wiki_qwen} report per-category Contain-Match Accuracy and LLM-Evaluation Accuracy (\%) for {\methodname} with GPT-4o-mini on the full 1{,}000-question evaluation sets, under two embedding backbones: all-MiniLM-L6-v2 and Qwen3-Embedding-0.6B.


\begin{table}[h]
\centering
\caption{Per-category accuracy on \textbf{MuSiQue} by hop count (GPT-4o-mini + all-MiniLM-L6-v2, $n{=}1{,}000$).}
\label{tab:musique_minilm}
\small
\begin{tabular}{lrcc}
\toprule
\textbf{Category} & \textbf{N} & \textbf{Contain Acc (\%)} & \textbf{LLM Acc (\%)} \\
\midrule
2-hop       & 518   & 57.7 & 56.8 \\
3-hop       & 316   & 44.0 & 50.6 \\
4-hop       & 166   & 31.9 & 28.9 \\
\cdashline{1-4}
\textit{Overall} & 1{,}000 & \textbf{49.1} & \textbf{50.2} \\
\bottomrule
\end{tabular}
\end{table}

\begin{table}[h]
\centering
\caption{Per-category accuracy on \textbf{HotpotQA} by question type (GPT-4o-mini + all-MiniLM-L6-v2, $n{=}1{,}000$).}
\label{tab:hotpotqa_minilm}
\small
\begin{tabular}{lrcc}
\toprule
\textbf{Category} & \textbf{N} & \textbf{Contain Acc (\%)} & \textbf{LLM Acc (\%)} \\
\midrule
Bridge      & 811   & 74.8 & 78.1 \\
Comparison  & 189   & 86.2 & 87.3 \\
\cdashline{1-4}
\textit{Overall} & 1{,}000 & \textbf{77.0} & \textbf{79.8} \\
\bottomrule
\end{tabular}
\end{table}

\begin{table}[h]
\centering
\caption{Per-category accuracy on \textbf{2WikiMultiHopQA} by question type (GPT-4o-mini + all-MiniLM-L6-v2, $n{=}1{,}000$).}
\label{tab:2wiki_minilm}
\small
\begin{tabular}{lrcc}
\toprule
\textbf{Category} & \textbf{N} & \textbf{Contain Acc (\%)} & \textbf{LLM Acc (\%)} \\
\midrule
Compositional      & 413   & 63.5 & 66.7 \\
Comparison         & 244   & 84.4 & 88.1 \\
Bridge-Comparison  & 235   & 89.2 & 85.6 \\
Inference          & 108   & 49.1 & 48.1 \\
\cdashline{1-4}
\textit{Overall}   & 1{,}000 & \textbf{73.0} & \textbf{74.3} \\
\bottomrule
\end{tabular}
\end{table}


\begin{table}[h]
\centering
\caption{Per-category accuracy on \textbf{MuSiQue} by hop count (GPT-4o-mini + Qwen3-Embedding-0.6B, $n{=}1{,}000$).}
\label{tab:musique_qwen}
\small
\begin{tabular}{lrcc}
\toprule
\textbf{Category} & \textbf{N} & \textbf{Contain Acc (\%)} & \textbf{LLM Acc (\%)} \\
\midrule
2-hop       & 518   & 50.0 & 51.0 \\
3-hop       & 316   & 39.6 & 40.2 \\
4-hop       & 166   & 24.7 & 22.3 \\
\cdashline{1-4}
\textit{Overall} & 1{,}000 & \textbf{42.5} & \textbf{42.8} \\
\bottomrule
\end{tabular}
\end{table}

\begin{table}[h]
\centering
\caption{Per-category accuracy on \textbf{HotpotQA} by question type (GPT-4o-mini + Qwen3-Embedding-0.6B, $n{=}1{,}000$).}
\label{tab:hotpotqa_qwen}
\small
\begin{tabular}{lrcc}
\toprule
\textbf{Category} & \textbf{N} & \textbf{Contain Acc (\%)} & \textbf{LLM Acc (\%)} \\
\midrule
Bridge      & 811   & 73.9 & 78.2 \\
Comparison  & 189   & 83.6 & 87.8 \\
\cdashline{1-4}
\textit{Overall} & 1{,}000 & \textbf{75.7} & \textbf{80.0} \\
\bottomrule
\end{tabular}
\end{table}

\begin{table}[h]
\centering
\caption{Per-category accuracy on \textbf{2WikiMultiHopQA} by question type (GPT-4o-mini + Qwen3-Embedding-0.6B, $n{=}1{,}000$).}
\label{tab:2wiki_qwen}
\small
\begin{tabular}{lrcc}
\toprule
\textbf{Category} & \textbf{N} & \textbf{Contain Acc (\%)} & \textbf{LLM Acc (\%)} \\
\midrule
Compositional      & 413   & 52.8 & 55.2 \\
Comparison         & 244   & 82.8 & 90.2 \\
Bridge-Comparison  & 235   & 88.1 & 75.7 \\
Inference          & 108   & 50.0 & 52.8 \\
\cdashline{1-4}
\textit{Overall}   & 1{,}000 & \textbf{68.1} & \textbf{68.3} \\
\bottomrule
\end{tabular}
\end{table}

\begin{figure*}[!t]
    \centering
    \includegraphics[width=0.9\linewidth]{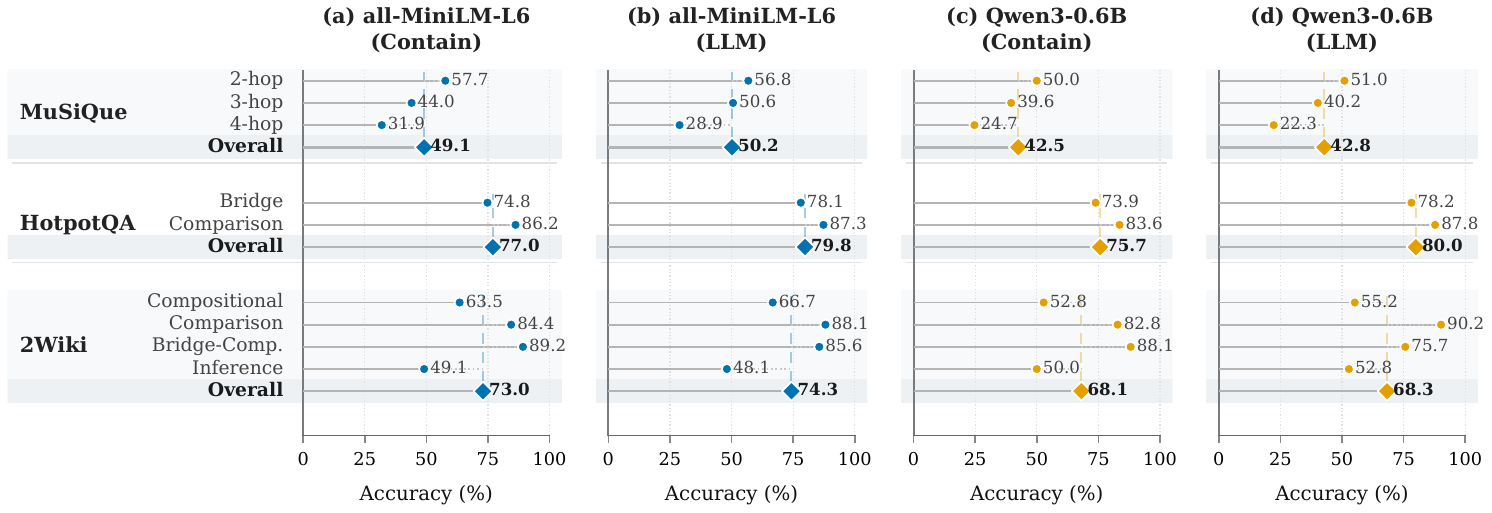}
    \Description{Per-category accuracy of ExecutionRAG under both embedding backbones: MuSiQue by hop count, HotpotQA by question type, and 2WikiMultiHopQA by question type.}
    \caption{\textbf{Per-category performance of {\methodname} under both embedding backbones.} We report Contain-Match Accuracy and LLM-Evaluation Accuracy on MuSiQue (by hop count), HotpotQA (by question type), and 2WikiMultiHopQA (by question type).}
    \label{fig:category}
\end{figure*}

\section{Retrieval Efficiency Analysis}
\label{app:efficiency}

We analyze the retrieval efficiency of {\methodname} relative to the strongest agentic baseline A-RAG and to our ablation variants.
Unless otherwise specified, all main comparisons use GPT-4o-mini with all-MiniLM-L6-v2 on the full 1{,}000-question evaluation sets.

\begin{table}[h]
\centering
\caption{Retrieval efficiency comparison on three benchmarks ($n{=}1{,}000$, GPT-4o-mini + all-MiniLM-L6-v2). \textbf{Tokens}: average retrieved tokens per question; \textbf{Cost}: average API cost per question (USD).}
\label{tab:efficiency_main}
\small
\renewcommand{\arraystretch}{1.15}
\setlength{\tabcolsep}{3.5pt}
\begin{tabular}{ll cccc}
\toprule
\textbf{Dataset} & \textbf{Method} & \textbf{LLM Acc} & \textbf{Cont Acc} & \textbf{Tokens/Q} & \textbf{Cost/Q} \\
\midrule
\multirow{2}{*}{MuSiQue}
  & A-RAG  & 44.9 & 42.0 & 5{,}015 & \$0.0034 \\
  & \textbf{Ours} & \textbf{50.2} & \textbf{49.1} & 7{,}161 & \$0.0064 \\
\midrule
\multirow{2}{*}{HotpotQA}
  & A-RAG  & 78.6 & 77.3 & 2{,}860 & \$0.0016 \\
  & \textbf{Ours} & \textbf{79.8} & \textbf{77.0} & 3{,}373 & \$0.0033 \\
\midrule
\multirow{2}{*}{2Wiki}
  & A-RAG  & 59.8 & 66.7 & 3{,}992 & \$0.0027 \\
  & \textbf{Ours} & \textbf{74.3} & \textbf{73.0} & 3{,}966 & \$0.0039 \\
\bottomrule
\end{tabular}
\end{table}

\begin{table}[h]
\centering
\caption{Retrieval efficiency under ablation on MuSiQue ($n{=}1{,}000$, GPT-4o-mini + all-MiniLM-L6-v2). We report token consumption and cost alongside the accuracy numbers from \cref{tab:ablation}.}
\label{tab:ablation_efficiency}
\small
\renewcommand{\arraystretch}{1.15}
\setlength{\tabcolsep}{2.8pt}
\begin{tabular}{ll cc cc}
\toprule
\textbf{Dataset} & \textbf{Variant} & \textbf{LLM} & \textbf{Cont} & \textbf{Tokens/Q} & \textbf{Cost/Q} \\
\midrule
\multirow{3}{*}{MuSiQue}
  & \cellcolor{gray!8}\textbf{Full} & \cellcolor{gray!8}50.2 & \cellcolor{gray!8}49.1 & \cellcolor{gray!8}7{,}161 & \cellcolor{gray!8}\$0.0064 \\
  & w/o Bootstrap        & 48.7 & 46.2 & 5{,}962 & \$0.0047 \\
  & w/o Progress Register & 48.3 & 48.5 & 6{,}378 & \$0.0042 \\
\bottomrule
\end{tabular}
\end{table}

\begin{table}[h]
\centering
\caption{Efficiency comparison across embedding models ($n{=}1{,}000$, GPT-4o-mini). {\methodname} maintains its accuracy advantage with both embedding backbones, with comparable or fewer retrieved tokens on 2WikiMultiHopQA.}
\label{tab:efficiency_embedding}
\small
\renewcommand{\arraystretch}{1.15}
\setlength{\tabcolsep}{2.8pt}
\begin{tabular}{ll ccc ccc}
\toprule
 & & \multicolumn{3}{c}{\textbf{all-MiniLM-L6-v2}} & \multicolumn{3}{c}{\textbf{Qwen3-Embed-0.6B}} \\
\cmidrule(lr){3-5} \cmidrule(lr){6-8}
\textbf{Dataset} & \textbf{Method} & LLM & Tok/Q & Cost/Q & LLM & Tok/Q & Cost/Q \\
\midrule
\multirow{2}{*}{MuSiQue}
  & A-RAG  & 44.9 & 5{,}015 & .0034 & 42.3 & 5{,}217 & .0040 \\
  & \textbf{Ours} & \textbf{50.2} & 7{,}161 & .0064 & \textbf{42.8} & 6{,}060 & .0083 \\
\midrule
\multirow{2}{*}{HotpotQA}
  & A-RAG  & 78.6 & 2{,}860 & .0016 & 76.4 & 3{,}533 & .0026 \\
  & \textbf{Ours} & \textbf{79.8} & 3{,}373 & .0033 & \textbf{80.0} & 3{,}568 & .0052 \\
\midrule
\multirow{2}{*}{2Wiki}
  & A-RAG  & 59.8 & 3{,}992 & .0027 & 55.2 & 4{,}311 & .0034 \\
  & \textbf{Ours} & \textbf{74.3} & 3{,}966 & .0039 & \textbf{68.3} & \textbf{4{,}044} & .0051 \\
\bottomrule
\end{tabular}
\end{table}

\begin{figure}[h]
    \centering
    \includegraphics[width=0.68\linewidth]{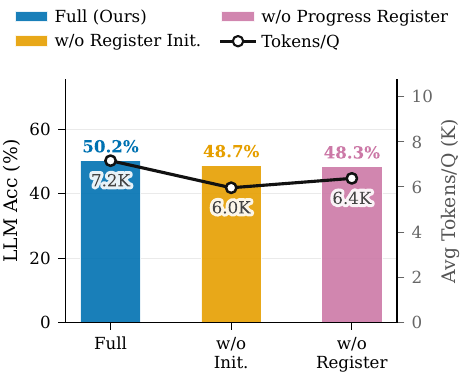}
    \Description{Accuracy against retrieved tokens per question for the ablation variants on MuSiQue; removing register components lowers tokens and accuracy together.}
    \caption{\textbf{Accuracy--efficiency trade-off across ablation
    variants} on MuSiQue, the same variants as Table~\ref{tab:ablation}.
    Removing a register component lowers retrieved tokens and accuracy
    together.}
    \label{fig:efficiency_ablation}
\end{figure}

\section{Failure Case Analysis}
\label{app:error}

To better understand the remaining failure modes of {\methodname}, we conduct a systematic error analysis on MuSiQue ($n{=}1{,}000$) and compare against the strongest agentic baseline, A-RAG. We report results under both embedding backbones to assess whether the observed patterns are robust to retriever choice.

\paragraph{Annotation method}
We leverage MuSiQue's ground-truth annotations, which provide for each question a set of \emph{supporting paragraphs} corresponding to the minimal evidence needed for a correct answer. We map each supporting paragraph to its corresponding corpus chunk using 40-character fingerprint matching (paragraph-level recall: 87.9\%), and compute \emph{hop-level evidence recall} as the fraction of supporting paragraphs for which at least one matching chunk was retrieved. Each failure ($\text{LLM-Acc}{=}0$) is assigned to exactly one category using the following priority order:

\begin{enumerate}[nosep,leftmargin=*]
  \item \textbf{Judge Error.} $\text{Contain-Acc}{=}1$ but $\text{LLM-Acc}{=}0$, indicating an evaluation mismatch.
  \item \textbf{Gave Up.} The predicted answer contains explicit abstention phrases (e.g., ``unable to find'', ``not found'').
  \item \textbf{Reasoning Error.} Either (a)~all supporting chunks were retrieved but the answer is still wrong, or (b)~[{\methodname} only] the register marks the question sufficiently answered despite remaining evidence gaps (\emph{false resolution}).
  \item \textbf{Missing Evidence.} Some supporting evidence was never retrieved and no false resolution occurred. This observable category includes both unreachable and unasked evidence.
\end{enumerate}

\paragraph{Reading the counts}
All comparisons below are reported as within-method percentages of each method's own total failures.
The totals themselves come from this failure-analysis pass, and for {\methodname} under all-MiniLM-L6-v2 the total is slightly larger than the failure count implied by that method's main-table accuracy, so the reduction reported for that pairing is if anything conservative.

\paragraph{Results (all-MiniLM-L6-v2)}

\begin{table}[h]
\centering
\caption{Failure case analysis on MuSiQue ($n{=}1{,}000$, GPT-4o-mini + \textbf{all-MiniLM-L6-v2}). Percentages are relative to total failures for each method.}
\label{tab:error_analysis}
\small
\renewcommand{\arraystretch}{1.15}
\setlength{\tabcolsep}{4pt}
\begin{tabular}{l cc cc}
\toprule
 & \multicolumn{2}{c}{\textbf{A-RAG}} & \multicolumn{2}{c}{\textbf{\methodname}} \\
\cmidrule(lr){2-3} \cmidrule(lr){4-5}
\textbf{Category} & \textbf{Count} & \textbf{\%} & \textbf{Count} & \textbf{\%} \\
\midrule
Total failures & 551 & 100.0 & 533 & 100.0 \\
\midrule
Judge Error      &  81 & 14.7 &  91 & 17.1 \\
Gave Up          &  36 &  6.5 &  70 & 13.1 \\
Reasoning Error  &  52 &  9.4 & 138 & 25.9 \\
Missing Evidence & 382 & \textbf{69.3} & 234 & \textbf{43.9} \\
\bottomrule
\end{tabular}
\end{table}

\begin{table}[h]
\centering
\caption{Failure distribution by hop count (\%, MiniLM). {\methodname} reduces Missing Evidence at every hop level ($-25.4$pp overall).}
\label{tab:error_by_hop}
\small
\renewcommand{\arraystretch}{1.15}
\setlength{\tabcolsep}{3.5pt}
\begin{tabular}{cl rrrr}
\toprule
\textbf{Hop} & \textbf{Method} & \textbf{Judge} & \textbf{Gave Up} & \textbf{Reasoning} & \textbf{Missing} \\
\midrule
\multirow{2}{*}{2-hop}
  & A-RAG         & 15.7 &  7.5 & 15.0 & \textbf{61.8} \\
  & {\methodname} & 18.6 & 14.2 & 30.1 & \textbf{37.2} \\
\midrule
\multirow{2}{*}{3-hop}
  & A-RAG         & 15.8 &  4.8 &  6.1 & \textbf{73.3} \\
  & {\methodname} & 17.7 & 14.3 & 24.0 & \textbf{44.0} \\
\midrule
\multirow{2}{*}{4-hop}
  & A-RAG         & 10.9 &  6.7 &  1.7 & \textbf{80.7} \\
  & {\methodname} & 13.6 &  9.8 & 21.2 & \textbf{55.3} \\
\bottomrule
\end{tabular}
\end{table}

\paragraph{Results (Qwen3-Embedding-0.6B)}

\begin{table}[h]
\centering
\caption{Failure case analysis on MuSiQue ($n{=}1{,}000$, GPT-4o-mini + \textbf{Qwen3-Embedding-0.6B}). Percentages are relative to total failures for each method.}
\label{tab:error_analysis_qwen}
\small
\renewcommand{\arraystretch}{1.15}
\setlength{\tabcolsep}{4pt}
\begin{tabular}{l cc cc}
\toprule
 & \multicolumn{2}{c}{\textbf{A-RAG}} & \multicolumn{2}{c}{\textbf{\methodname}} \\
\cmidrule(lr){2-3} \cmidrule(lr){4-5}
\textbf{Category} & \textbf{Count} & \textbf{\%} & \textbf{Count} & \textbf{\%} \\
\midrule
Total failures & 577 & 100.0 & 572 & 100.0 \\
\midrule
Judge Error      &  80 & 13.9 &  78 & 13.6 \\
Gave Up          &  69 & 12.0 &  99 & 17.3 \\
Reasoning Error  &  45 &  7.8 & 121 & 21.2 \\
Missing Evidence & 383 & \textbf{66.4} & 274 & \textbf{47.9} \\
\bottomrule
\end{tabular}
\end{table}

\begin{table}[h]
\centering
\caption{Failure distribution by hop count (\%, Qwen). {\methodname} reduces Missing Evidence at every hop level ($-18.5$pp overall).}
\label{tab:error_by_hop_qwen}
\small
\renewcommand{\arraystretch}{1.15}
\setlength{\tabcolsep}{3.5pt}
\begin{tabular}{cl rrrr}
\toprule
\textbf{Hop} & \textbf{Method} & \textbf{Judge} & \textbf{Gave Up} & \textbf{Reasoning} & \textbf{Missing} \\
\midrule
\multirow{2}{*}{2-hop}
  & A-RAG         & 16.1 & 13.9 & 12.5 & \textbf{57.5} \\
  & {\methodname} & 16.9 & 17.7 & 25.2 & \textbf{40.2} \\
\midrule
\multirow{2}{*}{3-hop}
  & A-RAG         & 11.4 &  9.1 &  5.7 & \textbf{73.9} \\
  & {\methodname} & 14.3 & 15.3 & 17.5 & \textbf{52.9} \\
\midrule
\multirow{2}{*}{4-hop}
  & A-RAG         & 12.5 & 11.7 &  0.8 & \textbf{75.0} \\
  & {\methodname} &  6.2 & 19.4 & 18.6 & \textbf{55.8} \\
\bottomrule
\end{tabular}
\end{table}

\paragraph{Key findings}
\textbf{First, Missing Evidence is reduced under both embedding backbones.}
With all-MiniLM-L6-v2, {\methodname} reduces Missing Evidence from 382
to 234 cases (69.3\%$\to$43.9\% of each method's failures). With
Qwen3-Embedding-0.6B, it reduces the count from 383 to 274
(66.4\%$\to$47.9\%).
This suggests that keeping the open sub-questions on an explicit register improves evidence acquisition robustly across retrievers.

\textbf{Second, evidence recovery exposes a different remaining bottleneck.}
As Missing Evidence decreases, Reasoning Error becomes more prominent for {\methodname} (MiniLM: 9.4\%$\to$25.9\%; Qwen: 7.8\%$\to$21.2\%).
These cases include false resolution, where the register marks an item answered too optimistically, as well as cases where retrieved evidence is present but incorrectly synthesized.

\textbf{Third, abstention increases as the register makes control more explicit.}
The \textbf{Gave Up} rate rises from 6.5\% to 13.1\% under MiniLM and from 12.0\% to 17.3\% under Qwen, indicating that {\methodname} is more willing to stop when the tracked evidence remains insufficient, rather than forcing a weakly supported answer.

\textbf{Finally, long-hop retrieval remains the dominant bottleneck.}
For {\methodname}, Missing Evidence still increases with hop count, reaching 55.3\% on 4-hop MuSiQue under MiniLM.
This suggests that although the register substantially improves the search, acquiring long and weakly connected evidence chains remains the main frontier for further progress.

\section{Case Study: How the Register Guides the Search}
\label{app:case}

To give qualitative intuition for how the register keeps the search on target and heads off common failures, we show three representative examples from MuSiQue (GPT-4o-mini + all-MiniLM-L6-v2). The \texttt{[Progress]} annotations in the {\methodname} column are the delta messages the agent actually sees: only register changes that materially affect the search are surfaced. The three cases illustrate three effects: anchoring on the right entity (Case~1), completing a long chain via progress deltas (Case~2), and self-correction by dropping a wrong item (Case~3).

\subsection*{Case 1: Preventing Entity Confusion (2-hop, Success)}

\noindent\textbf{Question:} \emph{Who married the publisher of abolitionist newspaper The North Star?}\\
\textbf{Gold answer:} Helen Pitts Douglass

\begin{table*}[t]
\centering
\small
\renewcommand{\arraystretch}{1.15}
\setlength{\tabcolsep}{4pt}
\begin{tabular}{p{0.47\linewidth}|p{0.47\linewidth}}
\toprule
\textbf{\methodname} (2 loops, 1 chunk) & \textbf{A-RAG} (5 loops, 2 chunks) \\
\midrule
\textbf{Warm-up:} Initial evidence correctly identifies Frederick Douglass as the publisher. The question is decomposed into two requirements: &
\textbf{Loop 1:} \texttt{keyword\_search} for ``The North Star, publisher, married.'' A top result mentions \emph{William Cooper Nell}, who \emph{wrote for} The North Star but was not its publisher. \\
\cmidrule(lr){1-1}\cmidrule(lr){2-2}

\texttt{req\_0}: ``identify the relevant marriage relation of Frederick Douglass'' $\to$ \colorbox{yellow!30}{\texttt{active}} &
\textbf{Loop 2:} \texttt{read\_chunk(126)} --- William Cooper Nell biography; no answer found. \\
\texttt{req\_1}: ``identify Frederick Douglass's wife'' $\to$ \texttt{pending} & \\
\cmidrule(lr){1-1}\cmidrule(lr){2-2}

\textbf{Loop 1:} Retrieved chunk 127 containing Douglass biography and spouse information. &
\textbf{Loop 3:} \texttt{keyword\_search} for ``William Cooper Nell, married, wife.'' The search trajectory is now anchored on the wrong entity. \\
\cmidrule(lr){1-1}\cmidrule(lr){2-2}

\textbf{State update:} &
\textbf{Loop 4:} Reads another irrelevant marriage-related chunk. \\
\texttt{req\_0}: remains \texttt{active} \\
\texttt{req\_1}: \texttt{pending} $\to$ \colorbox{green!25}{\texttt{resolved}} \\
Evidence: ``Helen Pitts Douglass, his second wife'' & \\
\cmidrule(lr){1-1}\cmidrule(lr){2-2}

\textbf{Answer:} \textcolor{teal}{\ding{51}} Helen Pitts Douglass &
\textbf{Loop 5:} Generates an answer based on the incorrect Nell branch. \\
&
\textbf{Answer:} \textcolor{red}{\ding{55}} William Cooper Nell / Sarah Jane Emery \\
\bottomrule
\end{tabular}
\caption{Case 1: {\methodname}'s seed-grounded register anchors the correct entity, Frederick Douglass, before iterative retrieval begins. A-RAG instead latches onto William Cooper Nell, a contributor rather than the publisher, and continues retrieval along the wrong branch.}
\label{tab:case1}
\end{table*}

\noindent\textbf{Analysis.}
This example illustrates the value of \emph{anchoring on the right entity} at the start. The seed-grounded register records Frederick Douglass as the entity the answer chain must stay attached to, so the later search stays on the correct branch. A-RAG has no explicit record of which entity the question has already grounded; once it latches onto Nell, later retrieval follows that mistaken branch and never recovers.

\subsection*{Case 2: Completing a Long Dependency Chain (4-hop, Success)}

\noindent\textbf{Question:} \emph{What day is the feast held, in the city that holds the headquarters of the only music company that is larger than the record label of Vilaiyaadu Mankatha?}\\
\textbf{Gold answer:} May 4

\begin{table*}[t]
\centering
\small
\renewcommand{\arraystretch}{1.15}
\setlength{\tabcolsep}{4pt}
\begin{tabular}{p{0.47\linewidth}|p{0.47\linewidth}}
\toprule
\textbf{\methodname} (6 loops, 3 chunks) & \textbf{A-RAG} (13 loops, 19 chunks) \\
\midrule
\textbf{Warm-up:} The question is decomposed into a dependency chain: &
\textbf{Loop 1:} Parallel searches for ``Vilaiyaadu Mankatha + music company + headquarters'' and ``feast + city + day.'' The latter is issued before the relevant city has been identified. \\
\texttt{req\_2}: ``identify the company larger than Sony Music'' $\to$ \texttt{pending} & \\
\texttt{req\_1}: ``identify the headquarters city of that company'' $\to$ \texttt{pending} & \\
\texttt{req\_0}: ``identify the feast day associated with that city'' $\to$ \colorbox{yellow!30}{\texttt{active}} & \textbf{Loops 2--3:} Reads multiple generic ``feast/city'' chunks, none linked to the music-company chain. \\
\cmidrule(lr){1-1}\cmidrule(lr){2-2}

\textbf{Loops 1--2:} \texttt{keyword\_search}(``Vilaiyaadu Mankatha, record label''); \texttt{read\_chunk(18,17)}. Chunk 18 resolves two links at once: Sony Music $\rightarrow$ Universal Music Group $\rightarrow$ Santa Monica. &
\textbf{Loops 4--6:} Repeated searches for ``largest music company, Sony Music, headquarters,'' with re-reading of overlapping evidence and no explicit record of intermediate progress. \\
\cmidrule(lr){1-1}\cmidrule(lr){2-2}

\textbf{State update (delta):} &
\textbf{Loop 7:} Eventually reads the key chunk, but only after substantial context has already been consumed by unrelated evidence. \\
\texttt{req\_2}: \texttt{pending} $\to$ \colorbox{green!25}{\texttt{resolved}} ``Universal Music Group'' \\
\texttt{req\_1}: \texttt{pending} $\to$ \colorbox{green!25}{\texttt{resolved}} ``Santa Monica, California'' & \\
\cmidrule(lr){1-1}\cmidrule(lr){2-2}

\textbf{Loops 3--5:} Conditioned on the updated state, the agent issues a targeted follow-up: \texttt{keyword\_search}(``feast, Santa Monica, California'') and \texttt{semantic\_search}(``feast Santa Monica California''). Reads chunk 19. &
\textbf{Loops 8--12:} Additional weakly constrained searches for ``feast day city'' continue without converging on the now-identified city. \\
\cmidrule(lr){1-1}\cmidrule(lr){2-2}

\textbf{Loop 6 (closure):} \texttt{req\_0} resolves with Saint Monica's feast day. &
\textbf{Loop 13:} Near budget exhaustion, the agent states that the feast is not explicitly mentioned and falls back to an incorrect alternative. \\
\textbf{Answer:} \textcolor{teal}{\ding{51}} May 4 &
\textbf{Answer:} \textcolor{red}{\ding{55}} ``Not explicitly mentioned'' \\
\bottomrule
\end{tabular}
\caption{Case 2: {\methodname} resolves a long dependency chain by turning register updates into targeted follow-up retrieval. Once the register records ``Universal Music Group'' and ``Santa Monica,'' the next search becomes sharply focused. A-RAG lacks an explicit record of this intermediate progress and keeps issuing weakly constrained searches.}
\label{tab:case2}
\end{table*}

\noindent\textbf{Analysis.}
This case shows how register updates turn a long multi-hop problem into a sequence of targeted retrieval steps. The key advantage is not decomposition alone, but the fact that newly answered items are surfaced as \emph{progress deltas}, which immediately sharpen the next search. Once ``Santa Monica'' is recorded on the register, the next retrieval step directly targets the remaining gap. Without such a register, A-RAG repeatedly searches in a much flatter space and accumulates substantial irrelevant context before the decisive entity is stabilized.

\subsection*{Case 3: Self-Correction via Requirement Retirement (4-hop, Success)}

\noindent\textbf{Question:} \emph{When did the birthplace of Ah Via Musicom's performer become the capital of the state Knowles is from?}\\
\textbf{Gold answer:} 1839

\begin{table*}[t]
\centering
\small
\renewcommand{\arraystretch}{1.15}
\setlength{\tabcolsep}{4pt}
\begin{tabular}{p{0.47\linewidth}|p{0.47\linewidth}}
\toprule
\textbf{\methodname} (6 loops, 3 chunks) & \textbf{A-RAG} (3 loops, 3 chunks) \\
\midrule
\textbf{Warm-up:} Initial evidence identifies Beyonc\'{e} Knowles as being from Texas, but incorrectly assumes that Houston is also the birthplace of the performer of \emph{Ah Via Musicom}. The initial requirement is therefore: &
\textbf{Loop 1:} Parallel searches for ``Ah Via Musicom, performer, birthplace'' and ``capital, state, Knowles.'' Both retrieve relevant but not yet integrated evidence. \\
\cmidrule(lr){1-1}\cmidrule(lr){2-2}

\texttt{req\_0}: ``determine when Houston became the capital of Texas'' $\to$ \colorbox{yellow!30}{\texttt{active}} &
\textbf{Loop 2:} Reads the relevant chunks identifying Eric Johnson and Texas, but does not revisit whether Houston is actually the correct city. \\
\cmidrule(lr){1-1}\cmidrule(lr){2-2}

\textbf{Loops 1--2:} \texttt{keyword\_search}(``Ah Via Musicom, performer, birthplace'') and \texttt{keyword\_search}(``capital, state, Knowles''). \texttt{read\_chunk(630,123)}. &
\textbf{Loop 3 (closure):} Produces a final answer based on an incorrect chain, conflating related Texas dates. \\
\cmidrule(lr){1-1}\cmidrule(lr){2-2}

\textbf{State update (loop 2):} The retrieved evidence contradicts the current requirement premise: Houston is not the capital of Texas, whereas Austin is. \\
\texttt{req\_0}: \texttt{active} $\to$ \colorbox{red!20}{\texttt{retired}} \\
Evidence: ``Houston is not the capital of Texas; Austin is the capital. Austin became the capital of the Republic of Texas in 1839.'' & \\
\cmidrule(lr){1-1}\cmidrule(lr){2-2}

\textbf{Loops 3--5:} After retiring the invalid requirement, {\methodname} issues targeted follow-ups: \texttt{keyword\_search}(``Eric Johnson, birthplace'') and \texttt{keyword\_search}(``Eric Johnson, Austin, Texas''). \texttt{read\_chunk(629)} confirms Eric Johnson was born in Austin. & \\
\cmidrule(lr){1-1}\cmidrule(lr){2-2}

\textbf{Loop 6 (closure):} The corrected chain is completed: Eric Johnson $\to$ Austin $\to$ Texas $\to$ 1839. &
\textbf{Answer:} \textcolor{red}{\ding{55}} 1836 \\
\textbf{Answer:} \textcolor{teal}{\ding{51}} 1839 & \\
\bottomrule
\end{tabular}
\caption{Case 3: {\methodname} uses the \texttt{retired} transition to explicitly discard an invalid requirement when later evidence contradicts its premise. A-RAG reads overlapping evidence but lacks an explicit mechanism for requirement revision, and therefore closes on a plausible but incorrect year.}
\label{tab:case3}
\end{table*}

\noindent\textbf{Analysis.}
This example highlights \emph{self-correction by dropping a wrong item}. Both systems initially move toward the wrong city, but {\methodname} can explicitly retire the invalid item once later evidence contradicts it. Dropping it is recorded as an explicit event rather than silently overwritten, which forces the search to reopen the chain with a corrected target. A-RAG has no explicit way to invalidate an earlier intermediate commitment, so its trajectory closes on a plausible but unsupported date.

\paragraph{Summary}
These three cases highlight the main qualitative effects of the register:
\begin{itemize}[nosep,leftmargin=*]
  \item \textbf{Anchoring on the right entity} (Case 1): building the register from seed evidence keeps retrieval on the correct entity and prevents early drift onto a wrong branch.
  \item \textbf{Delta-guided chain completion} (Case 2): answered items are surfaced as progress deltas, enabling sharply targeted follow-up retrieval instead of repeated weakly constrained search.
  \item \textbf{Self-correction by dropping a wrong item} (Case 3): contradictory evidence can explicitly invalidate an earlier item, letting the agent revise the chain instead of closing on a stale assumption.
\end{itemize}

\section{Prompt Templates}
\label{app:prompts}

This section lists the key prompts used in {\methodname}. All prompts are invoked with GPT-4o-mini unless otherwise stated.

\subsection*{A.1\quad Default QA Agent System Prompt}

Used to initialize the tool-using retrieval agent. The agent iteratively calls \texttt{keyword\_search}, \texttt{semantic\_search}, and \texttt{read\_chunk} while using the register as guidance.

\begin{table}[h]
\centering
\small
\setlength{\tabcolsep}{4pt}
\renewcommand{\arraystretch}{1.3}
\begin{tabular}{p{0.95\linewidth}}
\toprule
\textbf{Default QA Agent System Prompt} \\
\midrule
You are a question-answering assistant with access to a document corpus through available tools. \\[2pt]
Your goal is to answer questions accurately by finding and analyzing relevant information from the provided documents. \\[2pt]
\textbf{Available Tools} \\
\textbullet~\texttt{keyword\_search}: Find chunks by exact keyword matching \\
\textbullet~\texttt{semantic\_search}: Find chunks by semantic similarity \\
\textbullet~\texttt{read\_chunk}: Read the full content of a specific chunk \\[2pt]
\textbf{Strategy} \\
Work iteratively: search $\to$ read $\to$ evaluate $\to$ update progress $\to \cdots \to$ answer. For multi-hop questions, use the requirement state to keep track of what has already been established, what remains unresolved, and what should be searched next. \\[2pt]
\textbf{When Answering} \\
Ground your response in the retrieved documents. Cite the specific chunks that support your answer. Provide a clear, direct answer supported by evidence. If the evidence is still insufficient, state the remaining uncertainty rather than speculating beyond the documents. \\
\bottomrule
\end{tabular}
\end{table}

\subsection*{A.2\quad Seed-Grounded Requirement Initialization Prompt}

Run once before iterative retrieval begins. Given the question and seed evidence, this prompt builds the initial register (stored as the \texttt{requirement\_register}).

\begin{table}[h]
\centering
\small
\setlength{\tabcolsep}{4pt}
\renewcommand{\arraystretch}{1.3}
\begin{tabular}{p{0.95\linewidth}}
\toprule
\textbf{Seed-Grounded Requirement Initialization Prompt} \\
\midrule
\texttt{Question: \{query\}} \\[2pt]
\texttt{Available seed evidence:} \\
\texttt{\{info\_summary\}} \\[4pt]
Analyze the question and seed evidence, then output a JSON object with the following fields: \\[2pt]
\textbullet~\texttt{"evidence\_summary"}: concise summary of what is already known from the seed evidence. Preserve specific facts, names, dates, and numbers. \\
\textbullet~\texttt{"requirement\_register"}: a dictionary mapping \texttt{req\_id} to \texttt{\{"text": str, "status": "pending"|"active", "evidence\_notes": str\}}. Each item should represent a concrete information requirement needed to answer the question. Exactly one requirement should be marked \texttt{"active"}; the others should be \texttt{"pending"}. \\
\textbullet~\texttt{"reasoning\_note"}: one short sentence explaining why the active requirement is the best current target. \\[2pt]
Requirements should be concrete, searchable, and grounded in the seed evidence rather than speculative. \\[2pt]
Output ONLY the JSON object. \\
\bottomrule
\end{tabular}
\end{table}

\subsection*{A.3\quad Requirement-State Injection}

Appended to the agent context after warm-up. It exposes the current register as guidance without overriding the agent's autonomy.

\begin{table}[h]
\centering
\small
\setlength{\tabcolsep}{4pt}
\renewcommand{\arraystretch}{1.3}
\begin{tabular}{p{0.95\linewidth}}
\toprule
\textbf{Requirement-State Injection} \\
\midrule
\textbf{\#\# Current Requirement State} \\[2pt]
The following answer requirements have been identified: \\[2pt]
\texttt{\{requirement\_register\}} \\[4pt]
Use this state to decide what to search for next. Focus on unresolved requirements, preserve the evidence already gathered, and avoid repeating searches that do not advance the state. You may receive \texttt{[Progress]} messages describing requirement-state changes after new evidence is read. \\
\bottomrule
\end{tabular}
\end{table}

\subsection*{A.4\quad Requirement Register Update Prompt}

Called after each retrieval loop to update the register. The LLM decides whether new evidence advances, resolves, or invalidates each tracked item.

\begin{table}[h]
\centering
\small
\setlength{\tabcolsep}{4pt}
\renewcommand{\arraystretch}{1.3}
\begin{tabular}{p{0.95\linewidth}}
\toprule
\textbf{Requirement Register Update Prompt} \\
\midrule
\texttt{Question: \{query\}} \\[2pt]
\texttt{Current requirement register:} \\
\texttt{\{requirement\_register\}} \\[2pt]
\texttt{Newly retrieved information:} \\
\texttt{\{new\_evidence\}} \\[4pt]
Update the requirement register based on the new information. Output a JSON object with the following field: \\[2pt]
\textbullet~\texttt{"requirement\_register"}: dictionary mapping \texttt{req\_id} to \texttt{\{"text": str, "status": "pending"|"active"|"resolved"|"retired", "evidence\_notes": str\}} \\[2pt]
Update each requirement as follows: \\
\textbullet~Set \texttt{"resolved"} only if the currently available evidence is sufficient to support that requirement. \\
\textbullet~Set \texttt{"active"} for the requirement that should currently guide the next retrieval step. \\
\textbullet~Set \texttt{"retired"} only if the requirement is no longer valid or is contradicted by the new evidence. \\
\textbullet~Otherwise keep it \texttt{"pending"}. \\[2pt]
Keep ALL existing requirements; do not drop any. Update \texttt{evidence\_notes} with a concise summary of the supporting or contradicting facts currently available. \\[2pt]
Output ONLY the JSON object. \\
\bottomrule
\end{tabular}
\end{table}

\subsection*{A.5\quad Force Final Answer Prompt}

Appended when the agent reaches the maximum loop or token budget. It disables further tool calls and forces the agent to answer from the current evidence state.

\begin{table}[h]
\centering
\small
\setlength{\tabcolsep}{4pt}
\renewcommand{\arraystretch}{1.3}
\begin{tabular}{p{0.95\linewidth}}
\toprule
\textbf{Force Final Answer Prompt} \\
\midrule
You have reached the budget limit. You MUST now provide a final answer based on the information gathered so far. Do NOT call any more tools. If the evidence is sufficient, answer directly. If the evidence is still incomplete, provide the best bounded answer you can and clearly state the remaining uncertainty. \\
\bottomrule
\end{tabular}
\end{table}

\section{Broader Impacts}
\label{app:broader}

Our work improves multi-hop QA by keeping an explicit register of open
sub-questions during multi-step retrieval. Related language-model and
agent systems have been studied for autonomous scientific
discovery~\citep{wei2025ai}, few-shot tabular learning~\citep{yang2026forestllm},
paper-to-poster generation~\citep{zhang2025postergen}, and benchmarks that
probe the cognitive abilities such systems
require~\citep{zhou2025scientists}. These settings
illustrate applications in which an agent must maintain intermediate
requirements while integrating information from multiple sources. By
making evidence support and provenance explicit, {\methodname} can
improve transparency and auditability in such workflows.

However, the method still depends on the underlying retriever and LLM, so biases or errors in source documents may propagate. The framework could be misused in automated decision-making systems without human oversight, producing overconfident answers based on incomplete evidence. We recommend deployment in settings where accountability and verification are feasible.

\end{document}